\documentclass[10pt]{article}
\usepackage[preprint]{tmlr}
\usepackage[utf8]{inputenc}
\usepackage{url}
\usepackage{amsmath, amssymb}
\usepackage{booktabs}
\usepackage{graphicx}
\usepackage{enumitem}
\usepackage{array}
\usepackage{placeins}
\usepackage[hidelinks]{hyperref}
\newcommand{\MeanStd}[2]{\shortstack{#1\\$\pm$ #2}}
\title{Communication in modular robotic motor control: Bilateral controllers under realistic constraints}
\author{
  \name Jingwen Li \email jlii0556@student.monash.edu \\
  \addr Department of Data Science and AI, Monash University
  \AND
  \name Levin Kuhlmann \email levin@monash.edu.au \\
  \addr Department of Data Science and AI, Monash University
  \AND
  \name Jason Friedman \email jason@tau.ac.il \\
  \addr Department of Physical Therapy \& Sagol School of Neuroscience, Tel Aviv University
  \AND
  \name Gideon Kowadlo \email gideon@cerenaut.ai \\
  \addr Department of Data Science and AI, Monash University \\
  Cerenaut
}
\date{}

\begin{document}
\maketitle

\begin{abstract}
Robotic motor control in musculoskeletal systems requires accurate movement and robust postural stabilisation under signal-dependent noise (where noise magnitude scales with motor-command magnitude), while minimising energetic cost. Modular controller architectures offer one way to distribute these control demands across interacting modules, but it remains unclear how such organisation affects performance relative to monolithic control and how inter-module communication influences the resulting control solution. Motivated by bilateral organisation and interhemispheric communication in biological motor control, we introduce a recurrent controller comprising two GRU-based modules connected by a learnable, delayed inter-module communication channel and train it in a differentiable two-arm musculoskeletal simulator. We use success rate as the main performance measure, combining reaching success (final endpoint error $<2$ cm) and holding success (hold-phase variance $<2.5\times10^{-5}\,\mathrm{m}^2$). Across reaching and holding tasks, the modular architecture substantially outperforms a parameter-matched monolithic architecture, achieving a success rate of $87.6\% \pm 13.8\%$ compared with $68.9\% \pm 27.8\%$. Within the communicating bilateral controller, training with non-zero communication delays increased success rate by 11.3 percentage points and reduced energetic cost by approximately 17\% relative to zero-delay communication. Compared with the same bilateral architecture without inter-module communication, learned communication reduced endpoint and trajectory errors by approximately 27\% and 52\%, respectively, decreased movement variance by 13\%, and lowered CCI by 4\%. Our findings suggest that biologically inspired modular organisation can support robust movement under noise and energetic constraints, while inter-module communication and its timing influence trade-offs among precision, stability, and energetic cost.
\end{abstract}

\section{Introduction}
Robotic motor control operates under real-world physical constraints rather than idealised, noise-free environments~\citep{nishikawaNeuromechanicsIntegrativeApproach2007,codolMotorNetPythonToolbox2024a}. Even basic actions, such as moving an arm from one location to another or holding it in place, require a controller to balance competing demands for accuracy, postural stability, and perturbation resistance under signal-dependent noise and energetic costs~\citep{flashCoordinationArmMovements1985,hoganImpedanceControlApproach1985,todorovOptimalFeedbackControl2002a}. These competing demands are further constrained by energetic pressure: real systems cannot rely on arbitrarily large motor commands because high actuation consumes energy, increases physical load, and can amplify signal-dependent variability~\citep{harrisSignaldependentNoiseDetermines1998a,attwellEnergyBudgetSignaling2001a,saxenaPerformanceLimitationsSensorimotor2020}. Modular neural architectures, in which control is distributed across interacting modules, offer one way to organise these control demands and have been studied through modular motor-control approaches~\citep{wolpertMultiplePairedForward1998,harunoMOSAICModelSensorimotor2001,michaelsGoaldrivenModularNeural2020}, mixture-of-experts methods~\citep{shazeer2017,fedusSwitchTransformersScaling2021}, and modular reinforcement learning approaches~\citep{wangCPGBasedHierarchicalLocomotion2021,heModularStrategyDistributed2023}. This raises a broader design question of how interacting modules should coordinate: through independent parallel operation, unconstrained information sharing, or some intermediate form of constrained communication.

Biological motor systems have, over evolution, converged on a particular solution to this problem~\citep{rogersEvolutionHemisphericSpecialization2000a,vallortigaraSurvivalAsymmetricalBrain2005a}. The cerebrocortical components of the mammalian motor system are organised within the two hemispheres of the brain and communicate across hemispheres through a dedicated channel called the corpus callosum, with interhemispheric transfer subject to conduction delays~\citep{meyerInhibitoryExcitatoryInterhemispheric1995a,serrienDynamicsHemisphericSpecialization2006a}. Behavioural and neurophysiological evidence suggests that this bilateral organisation is not incidental: the two hemispheres employ complementary control strategies~\citep{sainburgHandednessDifferentialSpecializations2005a,woytowiczHandednessResultsComplementary2018a}, with interhemispheric communication playing an active role in reducing interference between conflicting motor demands~\citep{serrienDynamicsHemisphericSpecialization2006a}. This makes asymmetry a meaningful measure, because communication may change not only overall task performance but also how control contributions are distributed across the two hemispheres~\citep{muthaEffectsBrainLateralization2012a,sainburgConvergentModelsHandedness2014a}. Recent theoretical work further suggests that bilateral organisation with constrained communication may arise naturally under the joint pressures of energetic cost, circuit reliability, and task complexity~\citep{seoaneOptimalityPressuresLateralization2023a}. These observations make biological motor systems a useful source of hypotheses for robotic controllers that must coordinate modular subsystems under realistic communication constraints.

This biological analogy is especially relevant for humanoid robots, whose human-like morphology may allow them to operate in everyday human environments, including homes, workplaces, and public spaces. Many recent robot-learning approaches address motor control through reinforcement learning, policy optimisation, and related empirical methods, emphasising task performance in locomotion and manipulation benchmarks~\citep{elguea-aguinacoReviewReinforcementLearning2023,tangDeepReinforcementLearning2025,radosavovicRealworldHumanoidLocomotion2024}. In parallel, modular control has emerged as an important design direction in robotics and learning-based control, where decomposition is used to simplify complex problems and improve adaptability~\citep{wangCPGBasedHierarchicalLocomotion2021}. This motivates two targeted architectural questions for robotic motor control. First, can modular structure itself improve performance under realistic constraints? Second, can communication between modules change the control strategies that support robust performance under changing task and constraint demands?

To study these questions, we use MotorNet~\citep{codolMotorNetPythonToolbox2024a}, a differentiable musculoskeletal simulator that captures key biomechanical features while allowing controlled comparisons between modular and monolithic controller architectures. We tested two hypotheses. First, that a modular controller architecture can improve control relative to a monolithic controller architecture matched in total recurrent hidden dimension. Second, that learned communication between modules can change the resulting control strategy and remains useful when communication signals are delayed. Our main contributions are as follows:
\begin{enumerate}
\item We show that the bilateral modular controller substantially outperforms a parameter-matched monolithic controller under motor-control constraints. This supports bilateral two-controller organisation as a useful design choice for two-arm robotic motor control when accurate movement, robust postural stabilisation, and energetic cost must be balanced.
\item We show that learned inter-module communication changes how the bilateral controller operates by altering its control strategy and shifting output contributions toward a more contralateral organisation. This suggests that, under different motor-control constraints, inter-module communication can support different strategies for negotiating trade-offs among task performance, energetic cost, and actuation strategy.
\item We show that inter-module communication delays improve control performance and energetic efficiency relative to zero-delay communication. This indicates that, although communication delays are generally treated as a control constraint, they do not necessarily degrade performance when two communicating controllers both contribute to the control computation and can even improve it.
\end{enumerate}

\section{Related work}
Studies of handedness, motor lateralisation, and bimanual coordination suggest that the two hemispheres make complementary contributions to trajectory control, stabilisation, and interlimb coordination, and that interhemispheric interaction is functionally important during movement \citep{sainburgHandednessDifferentialSpecializations2005a,muthaEffectsBrainLateralization2012a,woytowiczHandednessResultsComplementary2018a,meyerInhibitoryExcitatoryInterhemispheric1995a,serrienDynamicsHemisphericSpecialization2006a,liuzziCoordinationUncoupledBimanual2011}. Other computational motor-control frameworks offer a complementary perspective by treating movement as optimisation under body dynamics, prediction, feedback, and uncertainty \citep{flashCoordinationArmMovements1985,wolpertComputationalApproachesMotor1997,kawatoInternalModelsMotor1999,todorovOptimalFeedbackControl2002a}. Together, these lines of work motivate bilateral organisation as a useful substrate for embodied control, while recent robotics-facing discussions have begun to argue that it may also offer design principles for robot motor systems \citep{rinaldoLeftRightBrain2025,dexheimerRolesHandednessHemispheric2024}.

What remains less clear is how these ideas translate into learnable controller architectures. In robotics and machine learning, modular control is often used to improve task performance, adaptability, or coordination, including through reinforcement learning, structured motor-control approaches, or distributed embodied systems \citep{tangDeepReinforcementLearning2025,elguea-aguinacoReviewReinforcementLearning2023,rajeswaranLearningComplexDexterous2018,wolpertMultiplePairedForward1998,harunoMOSAICModelSensorimotor2001,michaelsGoaldrivenModularNeural2020,wangCPGBasedHierarchicalLocomotion2021,heModularStrategyDistributed2023}. In parallel, bilateral or lateralised controller architectures have been studied in language, vision, and other representational domains \citep{changUnifiedNeurocomputationalBilateral2020a,dobsBrainlikeFunctionalSpecialization2022a,rajagopalanDeepLearningBilateral2025a,behrmannEmergenceTopographyHemispheric2026}. However, these studies generally target task performance or representation rather than biomechanical motor control under matched constraints, and rarely separate the effect of a modular controller architecture from the effect of inter-module communication during task execution.

\section{Method}
\subsection{Musculoskeletal environment}
\begin{figure}[h]
\centering
\includegraphics[width=0.9\linewidth]{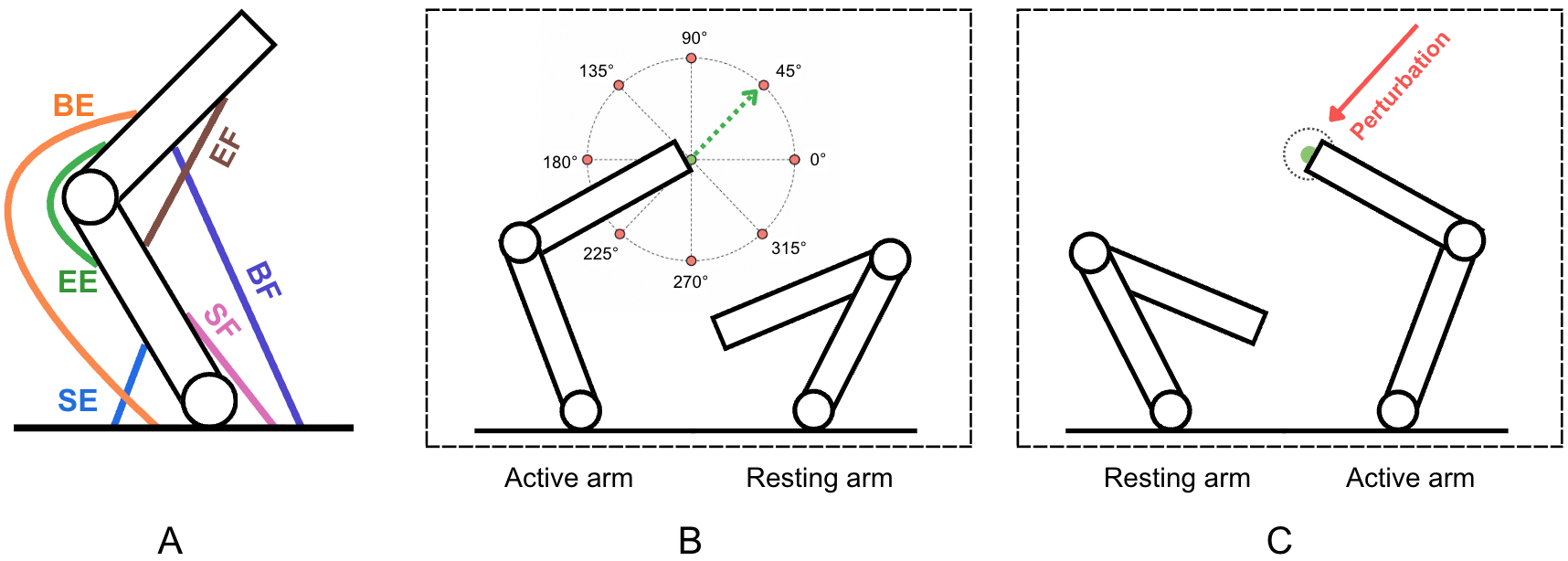}
\caption{Musculoskeletal substrate and task design. (A) A MotorNet RigidTendonArm26 effector: a planar two-degree-of-freedom arm actuated by six Hill-type muscles: shoulder flexor/extensor (SF/SE), elbow flexor/extensor (EF/EE), and biarticular flexor/extensor (BF/BE). The bimanual setup places two arms symmetrically in a shared workspace. (B) Centre-out reaching task: the active hand moves from the start position (green) to one of eight circularly arranged targets (red), while the resting hand remains undisturbed. (C) Postural holding task: the active hand maintains its position (green) against an external perturbation (red arrow) within a tolerance region (dashed circle), while the resting hand remains undisturbed.}
\label{fig:arm}
\end{figure}

We use MotorNet~\citep{codolMotorNetPythonToolbox2024a}, a differentiable musculoskeletal simulator that supports muscle-level actuation, differentiable task objectives, and access to task-relevant kinematic states needed for experimental manipulations and measurements central to our study. The bimanual system uses two RigidTendonArm26 effectors (Figure~\ref{fig:arm}A), each a planar two-degree-of-freedom arm actuated by six Hill-type muscles, giving twelve muscles in total. The two shoulders are placed symmetrically with a separation of 0.4 m.

\vspace{-2mm}

\subsection{Tasks}
\vspace{-1mm}
Motivated by classic paradigms in human motor-control research, we train and test the controllers on two complementary tasks that emphasise goal-directed reaching and postural stabilisation \citep{flashCoordinationArmMovements1985,hoganImpedanceControlApproach1985,sainburgHandednessDifferentialSpecializations2005a,woytowiczHandednessResultsComplementary2018a}. These tasks allow us to examine the controller architectures under distinct trajectory- and stability-related motor demands. Training follows an interleaved protocol that alternates between reaching and holding batches. In each batch of 128 trials, 64 use the left hand as active and 64 use the right hand, ensuring symmetric exposure. After one reaching batch and one holding batch, the two losses are normalised and summed for a single parameter update.

\begin{itemize}[leftmargin=1.5em]
    \item \textbf{Centre-out reaching task.} The active hand executes a centre-out movement from a central start position to one of eight targets evenly spaced on a $0.08\,\mathrm{m}$ radius circle, while the resting hand remains at its starting position. This layout samples eight movement directions and allows reaching performance to be evaluated across directions (Figure~\ref{fig:arm}B).
    \item \textbf{Postural holding task.} The active hand maintains its initial position while resisting a zero-mean Gaussian external force perturbation with a per-axis standard deviation of $0.5\,\mathrm{N}$, while the resting hand remains undisturbed. This task evaluates postural stability under external disturbances (Figure~\ref{fig:arm}C).
\end{itemize}

\subsection{Main controller architectures and baselines}
Inspired by the bilateral organisation of the brain, in which the two cerebral hemispheres exchange information through the corpus callosum~\citep{rogersEvolutionHemisphericSpecialization2000a,meyerInhibitoryExcitatoryInterhemispheric1995a,pauwelsRoleCorpusCallosum2023}, we evaluate three controller architectures. We use CC to denote the corpus-callosum-inspired inter-module communication channel in the bilateral controller. Each bilateral controller comprises a recurrent network with two interacting GRU-based modules. Each module contains a GRU-64 recurrent core and a module-specific readout branch. Bilateral w/o CC removes the communication channel between the two modules, allowing us to assess the contribution of inter-module communication. Monolithic replaces the two-module controller with a GRU-128 recurrent core and no modular decomposition, allowing us to assess the contribution of modular organisation.
\vspace{-4mm}

\begin{figure}[h]
\centering
\includegraphics[width=0.9\linewidth]{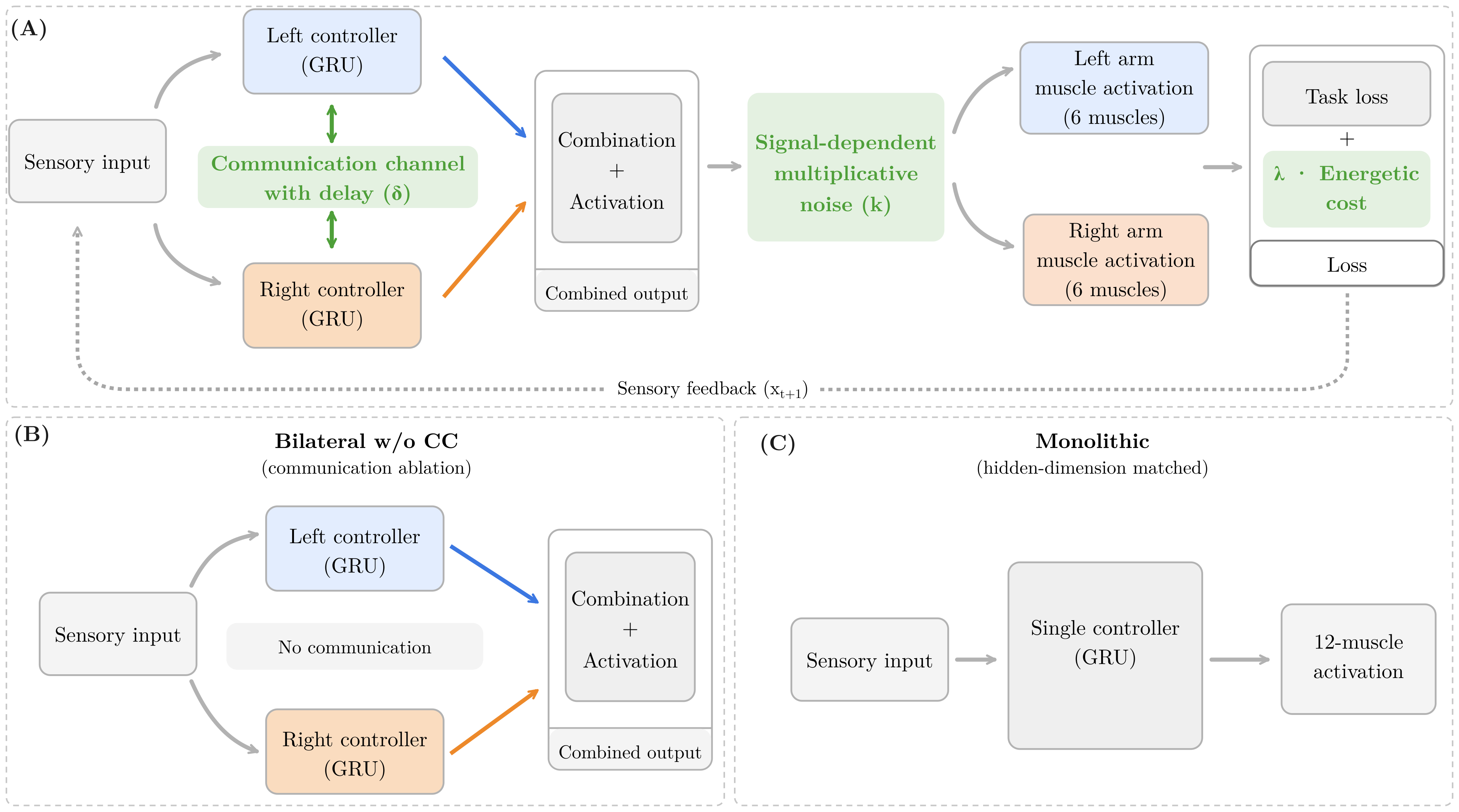}
\caption{Controller architectures. (A) Bilateral w/ CC consists of two GRU-based modules receiving shared sensory input and connected by a learnable bidirectional communication channel with delay $\delta$. Their outputs form a 12-dimensional muscle-command vector. The first six components actuate the left arm and the remaining six the right arm, while both modules contribute to all 12 commands. Signal-dependent motor noise is controlled by $k$ and energetic cost is weighted by $\lambda$. (B) Bilateral w/o CC retains the two-module structure but removes inter-module communication. (C) Monolithic replaces the bilateral decomposition with a single GRU-128 controller matched to the bilateral controller's total recurrent hidden dimension.}
\label{fig:architecture}
\end{figure}

\begin{itemize}[leftmargin=1.5em]
\item \textbf{Bilateral w/ CC (main controller architecture).} The controller comprises two GRU-based modules connected by a learnable bidirectional communication channel with delay $\delta$ (Figure~\ref{fig:architecture}A and Appendix~\ref{app:architecture}). Each module processes sensory input and delayed information from the other module, and the two module-specific readout contributions are combined to generate muscle activations for both arms. Both modules receive the same sensory observation and can contribute to all 12 muscle commands. No fixed lateralised or contralateral output routing is imposed.
\item \textbf{Bilateral w/o CC (communication-ablation).} The same two GRU-based modules and readout structure as Bilateral w/ CC, with the cross-module projections disabled (Figure~\ref{fig:architecture}B).
\item \textbf{Monolithic (hidden-dimension-matched).} A single GRU with hidden size 128 and no modular decomposition, matching the bilateral controller's total recurrent hidden dimension (Figure~\ref{fig:architecture}C).
\end{itemize}
\vspace{-2mm}

Because removing the cross-module projections also removes their parameters, we report two additional controller architectures as parameter-matching controls: (1) an exactly parameter-matched \textbf{self-projection architecture (SelfProj)} with the same parameter count as Bilateral w/ CC. In SelfProj, each cross-module projection is replaced by an equally sized projection of the same module's previous hidden state, preserving the parameter count without inter-module communication. (2) A near-parameter-matched \textbf{monolithic GRU-137 architecture (Monolithic-137)} has 78,513 parameters compared with 78,092 for Bilateral w/ CC, using the nearest larger integer hidden size to avoid disadvantaging the monolithic baseline. Full architectural definitions are provided in Appendix~\ref{app:architecture}. Parameter counts and the matching-control experiments are provided in Appendix~\ref{app:param_match}.

\subsection{Motor-control constraints}
\label{sec:constraints}
We evaluate the controllers under three biologically motivated motor-control constraints affecting complementary aspects of control. These constraints allow us to examine how inter-module communication changes as temporal, execution, and energetic pressures vary. For Bilateral w/ CC, these factors form 64 conditions. Architectures without communication are evaluated over the shared 16-condition noise--energy grid.

\textbf{Inter-module communication delay $\delta$.} Interhemispheric communication in biological motor systems is subject to non-zero conduction delays \citep{meyerInhibitoryExcitatoryInterhemispheric1995a,serrienDynamicsHemisphericSpecialization2006a}. Motivated by this temporal constraint, we introduce an explicit delay in the bidirectional inter-module communication channel. We evaluate $\delta \in \{0,5,10,20\}$ simulation steps, corresponding to 0, 50, 100, and 200 ms. For non-zero $\delta$, a ring buffer provides each module with past activity from the other module at the specified delay. The tested delays span progressively stronger temporal constraints and are intended as controlled computational manipulations instead of direct estimates of physiological conduction times.

\textbf{Signal-dependent motor noise $k$.} To evaluate controllers under more challenging motor-control conditions, we introduce signal-dependent motor noise motivated by human motor control, where movement variability increases with motor-command magnitude \citep{harrisSignaldependentNoiseDetermines1998a}. We model this effect using multiplicative noise applied to the motor commands, with noise magnitude controlled by $k$. We evaluate $k \in \{0,0.05,0.1,0.2\}$, where $k=0$ gives deterministic motor execution.

\textbf{Energetic-cost weight $\lambda$.} To evaluate whether control strategies change under pressure to reduce motor expenditure, we introduce an energetic cost motivated by biological motor control, where movement is influenced by energetic demands \citep{todorovOptimalFeedbackControl2002a}. We incorporate this pressure by adding an energetic-cost term to the training objective, weighted by $\lambda$. We evaluate $\lambda \in \{0,10^{-4},10^{-3},10^{-2}\}$, with $\lambda=0$ corresponding to no energetic penalty.

\subsection{Evaluation metrics}
We use success rate as the primary task-level performance metric for the reach--hold control problem. It combines reach success, defined as final reach error below $0.02\,\mathrm{m}$, and hold success, defined as hold-phase variance below $2.5\times10^{-5}\,\mathrm{m}^2$. The thresholds correspond to a 2 cm reach tolerance and $\sim$5 mm holding variability, consistent with human reaching and posture ranges~\citep{ongTargetSizeManipulations2019,berretReachEndpointFormation2014,shinoharaFluctuationsMotorOutput2008}. In addition to this combined metric, we also report endpoint error, trajectory error, movement variance, hold variance, energetic cost, and co-contraction index (CCI). Lower values indicate better precision or stability for the kinematic and variance-based metrics, and lower motor expenditure for energetic cost. CCI reflects the relative balance of antagonist activation and is interpreted together with energetic cost. Lower CCI alongside higher energetic cost is consistent with more selective antagonist recruitment and continued active control, whereas higher CCI at very low energetic cost can reflect weak but balanced co-activation. Asymmetry analyses use contralaterality index (CI) from output contributions and double-dissociation index (DDI) from lesion effects. Full definitions are in Appendix~\ref{app:metrics}.

\subsection{Implementation details}
Models from the three controller architectures were optimised on interleaved reaching and holding batches using Adam~\citep{kingmaAdamMethod2015} with a learning rate of $10^{-3}$ and batch size 128. Training used 10-step truncated backpropagation through time, gradient clipping, and early stopping. For the main comparison, all three architectures were evaluated over the $4\times4$ noise--energy grid defined in Section~\ref{sec:constraints}, with ten random seeds per condition. For Bilateral w/ CC, this grid was additionally evaluated across communication-delay settings. In the analysis, we aggregate the non-zero-delay conditions ($\delta\in\{5,10,20\}$) because our primary question is whether communication remains useful under delayed signalling. Using several delay magnitudes reduces dependence on any single delay condition, with delay-specific results reported in Appendix~\ref{app:delay}. All models were implemented in PyTorch~\citep{paszke2019pytorch} and trained on NVIDIA L40S GPUs.

The code snapshot is available at \url{https://anonymous.4open.science/r/motor-control-code-75F2/}.

\section{Experimental results}
\subsection{Modularity, the dominant source of performance gain under motor-control constraints}
\label{sec:modularity}
\begin{figure}[h]
\centering
\includegraphics[width=0.9\linewidth]{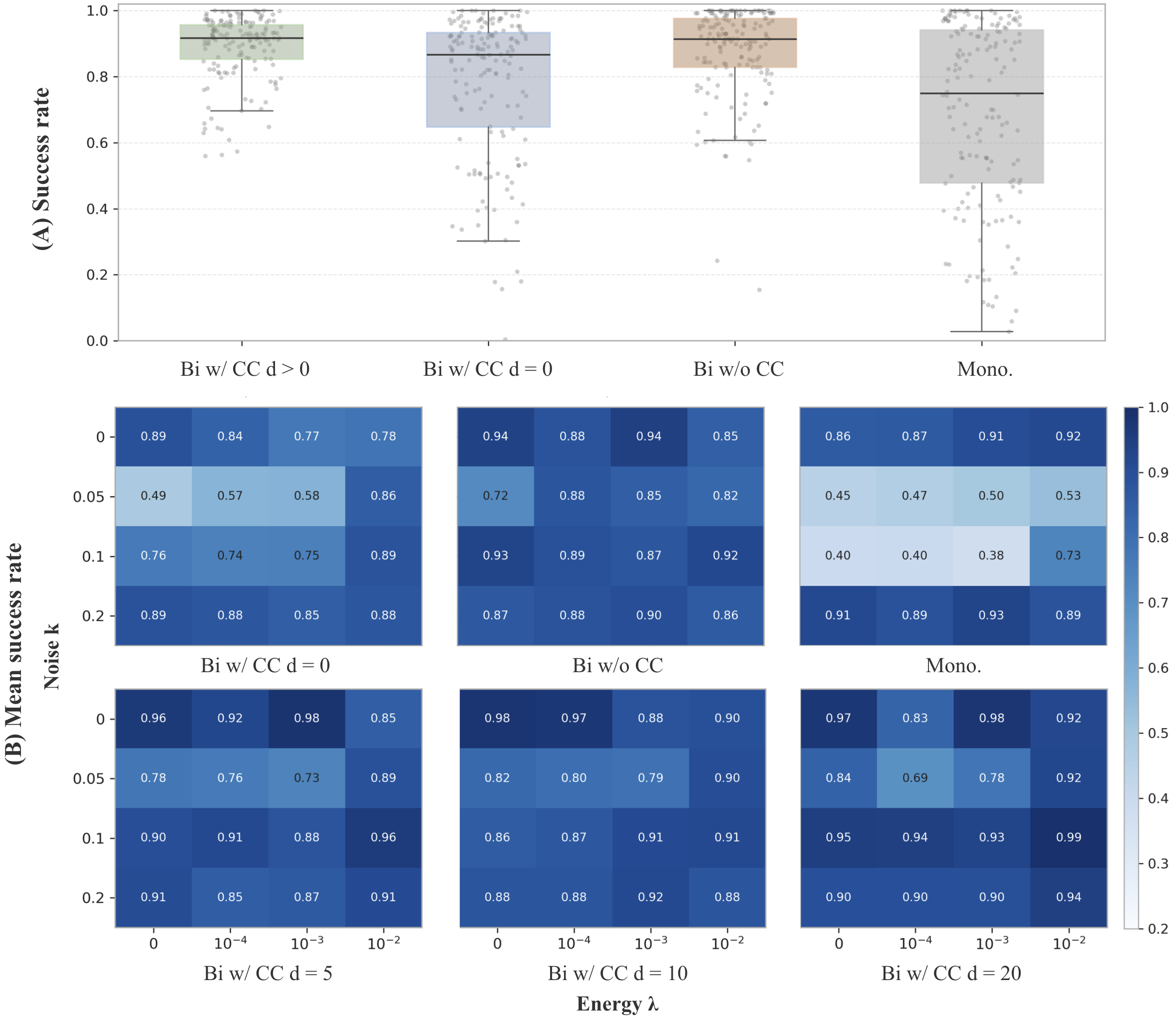}
\caption{Task-performance profiles across controller architectures and constraint settings. (A) Success-rate distributions across the noise--energy grid and ten seeds. Boxes show medians and interquartile ranges, whiskers show non-outlier spread, and points show configuration values. For Bilateral w/ CC at $\delta>0$, values are first averaged over $\delta\in\{5,10,20\}$ within each $(\lambda,k,\mathrm{seed})$ setting. (B) Mean success rate across $(\lambda,k)$ conditions, averaged over ten seeds after within-seed delay aggregation for $\delta>0$.}
\label{fig:modularity_success}
\end{figure}

Figure~\ref{fig:modularity_success}A shows whether modular controller architectures improve success rate under the shared noise--energy grid: the non-zero-delay Bilateral w/ CC conditions are strongest, Bilateral w/o CC is close behind, the zero-delay Bilateral w/ CC condition is weaker, and trained models using the modular controller architectures outperform those using the monolithic controller architecture. This modularity advantage is also evident in the exactly parameter-matched comparison between Bilateral w/o CC and Monolithic-128, which both contain 69,900 trainable parameters: Bilateral w/o CC achieves a substantially higher mean success rate (87.59\% vs.\ 68.91\%, +18.69 percentage points, $p<0.05$), showing that the task-success advantage is not explained by parameter count (Appendix~\ref{app:param_match}). Figure~\ref{fig:modularity_success}B resolves this result across motor-noise regimes. At low noise, trained models from all three primary controller architectures achieve high success rate, as expected when the task imposes little signal-dependent pressure. At intermediate noise ($k=0.05,0.1$), trained models using the monolithic controller architecture become fragile, with success rate falling into the 38.00\%--53.00\% range, whereas those using the modular controller architectures remain more robust.

The high-noise regime ($k=0.2$) is more nonlinear. Trained models using the modular controller architectures remain robust, but the monolithic baseline also shows an apparent rebound in success rate. This does not mean that the monolithic trained models have learned stronger active control. We interpret this cautiously: because motor noise is signal-dependent, low commanded activation also reduces injected noise and energetic cost, so a low-output trained model can satisfy the original hold criterion without providing stronger active stabilisation. Perturbation-strength diagnostics support this interpretation: when the holding perturbation is strengthened, the monolithic high-noise rebound becomes fragile (Appendix~\ref{app:high_noise}). Taken together, these regime-dependent patterns show that modularity improves overall task success and makes the trained models more robust across the tested motor-control constraints.

\subsection{Metric decomposition of the modularity gain}
\label{sec:modularity2}
Table~\ref{tab:main_results} decomposes the task-performance result from Figure~\ref{fig:modularity_success} into kinematic and actuation-level metrics. The Bilateral w/ CC, $\delta>0$ conditions provide the strongest profile, with the highest success rate and the best mean value on all decomposition metrics. In contrast, models using the monolithic controller architecture are significantly worse on every displayed metric ($p<0.05$, Appendix~\ref{app:statistics}), with lower success rate and higher kinematic, stability, and energetic costs. This supports the main modularity result: splitting a controller across two recurrent modules is the dominant source of the performance gain under the tested constraints.

\begin{table}[h]
\centering
\caption{Metric decomposition across matched energy--noise condition and seed combinations (ten seeds, four $\lambda$ values, four noise levels, $N=160$). Bilateral w/ CC ($\delta>0$) averages each metric over $\delta\in\{5,10,20\}$ within each matched energy--noise condition and seed before paired testing and is the reference column for all asterisks. Asterisks mark scores in the other columns that differ significantly from Bilateral w/ CC ($\delta>0$) under Bonferroni-corrected paired Wilcoxon tests ($p<0.05$, corrected over the three column-wise comparisons within each metric). Full test details are in Appendix~\ref{app:statistics}. Arrows indicate the direction of improvement ($\uparrow$ higher is better, $\downarrow$ lower is better), and bold marks the best mean in each row.}
\label{tab:main_results}
\begin{tabular*}{\textwidth}{@{\extracolsep{\fill}}lcccc@{}}
\toprule
Metric & Bi w/ CC, $\delta>0$ & Bi w/ CC, $\delta=0$ & Bi w/o CC & Mono. \\
\midrule
Success rate $\uparrow$ & \MeanStd{\textbf{88.81\%}}{10.01\%} & \MeanStd{77.53\%$^{*}$}{21.99\%} & \MeanStd{87.59\%}{13.82\%} & \MeanStd{68.91\%$^{*}$}{27.81\%} \\
Endpoint error $\downarrow$ & \MeanStd{\textbf{0.0085}}{0.0017} & \MeanStd{0.0091$^{*}$}{0.0022} & \MeanStd{0.0117$^{*}$}{0.0041} & \MeanStd{0.0091$^{*}$}{0.0023} \\
Trajectory error $\downarrow$ & \MeanStd{\textbf{0.0022}}{0.0003} & \MeanStd{0.0025$^{*}$}{0.0005} & \MeanStd{0.0046$^{*}$}{0.0025} & \MeanStd{0.0026$^{*}$}{0.0005} \\
Movement variance $\downarrow$ & \MeanStd{\textbf{0.0081}}{0.0013} & \MeanStd{0.0092$^{*}$}{0.0038} & \MeanStd{0.0093$^{*}$}{0.0027} & \MeanStd{0.0099$^{*}$}{0.0030} \\
Hold variance $\downarrow$ & \MeanStd{\textbf{1.054e{-}5}}{6.723e{-}6} & \MeanStd{1.556e{-}5$^{*}$}{1.471e{-}5} & \MeanStd{1.128e{-}5}{8.132e{-}6} & \MeanStd{1.867e{-}5$^{*}$}{1.338e{-}5} \\
Energetic cost $\downarrow$ & \MeanStd{\textbf{1.0875}}{1.6950} & \MeanStd{1.3196$^{*}$}{1.7233} & \MeanStd{1.1930}{1.8288} & \MeanStd{1.7666$^{*}$}{1.9085} \\
CCI $\downarrow$ & \MeanStd{\textbf{0.7175}}{0.0369} & \MeanStd{0.7340$^{*}$}{0.0401} & \MeanStd{0.7473$^{*}$}{0.0340} & \MeanStd{0.7606$^{*}$}{0.0466} \\
\bottomrule
\end{tabular*}
\end{table}

The modular ablations refine this result. Bilateral w/o CC remains statistically indistinguishable in success rate, showing that modular structure alone avoids much of the monolithic controller's fragility, while its endpoint error, trajectory error, movement variance, and CCI differ significantly from the non-zero-delay Bilateral w/ CC reference ($p<0.05$, Appendix~\ref{app:statistics}). The zero-delay Bilateral w/ CC condition is significantly worse than the non-zero-delay condition on every displayed metric. These differences among modular architectures and conditions motivate the next question: whether learned communication changes the operating regime of an already strong modular controller, beyond its effect on whether the task succeeds.

\subsection{Learned communication changes the modular control strategy}
\label{sec:communication}
The inter-module communication effect appears as a shift in operating regime. Table~\ref{tab:main_results} shows that, relative to the non-zero-delay Bilateral w/ CC reference, both zero-delay communication and the communication-free modular baseline occupy different points in the same trade-off space. Non-zero-delay Bilateral w/ CC has significantly lower CCI than Bilateral w/o CC, despite no significant difference in aggregate success, and combines significantly stronger success with lower energetic cost than the zero-delay Bilateral w/ CC condition. The advantage over zero delay is not driven by one selected delay: each of $\delta=5$, 10, and 20 significantly improves success and the four reported decomposition metrics after correction ($p<0.05$, Appendix~\ref{app:delay}).

\begin{figure}[h]
\centering
\includegraphics[width=0.99\linewidth]{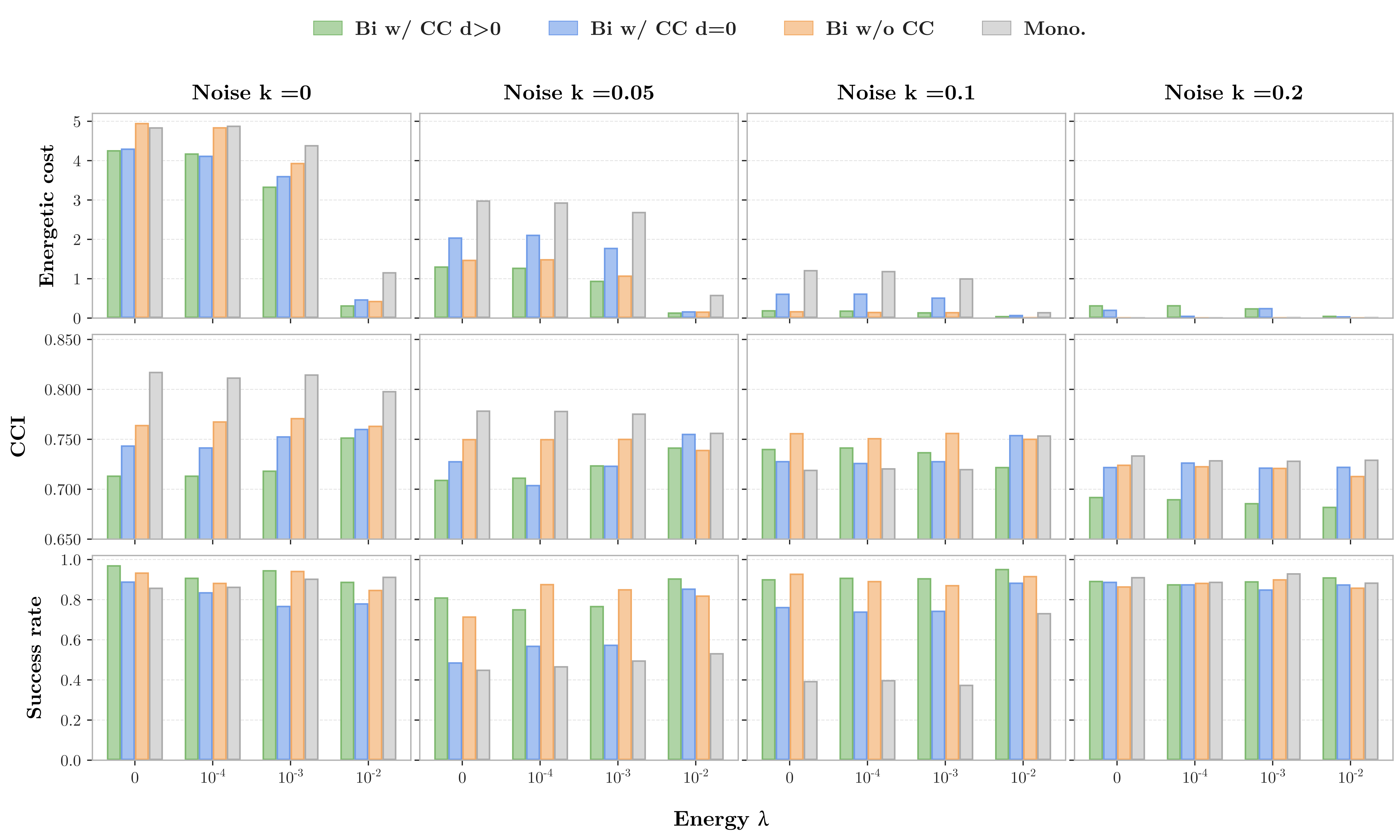}
\caption{Energetic cost, CCI, and success rate across constraint regimes. Panels within each row share a common y-axis scale. Each column corresponds to a noise condition ($k = 0, 0.05, 0.1, 0.2$). Within each panel, bars show four comparison groups (Bilateral w/ CC averaged over $\delta \in \{5,10,20\}$, Bilateral w/ CC at $\delta=0$, Bilateral w/o CC, and Monolithic) at each energy-penalty weight. Each bar shows the mean across ten random seeds for the corresponding $(\lambda,k)$ condition. For Bilateral w/ CC at $\delta>0$, values are first averaged over $\delta \in \{5,10,20\}$ within each seed and then averaged across seeds.}
\label{fig:regime_bars}
\end{figure}

Figure~\ref{fig:regime_bars} compares energetic cost, co-contraction, and success rate across noise and energy conditions. At low-to-moderate noise, non-zero-delay Bilateral w/ CC shows its clearest gains in success rate and energetic cost, alongside clearer separation between comparison groups in co-contraction and actuation strategy. The lower co-contraction index (CCI) provides an actuation-level signature of this shift, indicating more selective muscle recruitment with less broad antagonist co-activation. Together, these patterns indicate a favourable performance--energetic-cost trade-off: delayed inter-module communication helps the controller complete the task more reliably with lower energetic cost without simply increasing motor output. At high noise ($k=0.2$), success rates become more similar across the modular conditions, but energetic cost and CCI reveal distinct control strategies. Because the motor noise is signal-dependent, reducing muscle activation also reduces the magnitude of the injected variability, so very low energetic cost can reflect a low-output strategy and does not necessarily indicate more efficient active stabilisation. Bilateral w/ CC at $\delta=0$, Bilateral w/o CC, and Monolithic move toward low-energy regimes while retaining non-zero CCI, consistent with weak but relatively balanced antagonist activation. In contrast, Bilateral w/ CC at $\delta>0$ retains measurable energetic cost together with relatively low CCI, consistent with continued active control using more selective antagonist recruitment. Non-zero-delay communication therefore changes how the controller achieves success under strong signal-dependent noise, beyond changes in task performance alone.

\subsection{Extension to simultaneous bimanual control}
\label{sec:bimanual}
To examine whether the main findings extend beyond the interleaved training protocol, where only one arm is active at a time, we repeated the experiment under simultaneous bimanual control. The reaching and holding tasks were unchanged, but were performed simultaneously, with one hand executing the centre-out reaching task while the other maintained postural stability under perturbation. All other experimental settings were retained, and models were trained with five random seeds. For this simultaneous setting, success rate is measured directly as the proportion of trials in which both the reaching and holding criteria are satisfied within the same rollout.

As shown in Table~\ref{tab:simultaneous_bimanual}, the main findings extend to the more complex scenario where both arms must satisfy distinct motor demands concurrently. Bilateral w/ CC ($\delta>0$) achieves the highest success rate (83.48\%), compared with 78.41\% for Bilateral w/o CC, 66.15\% for Bilateral w/ CC ($\delta=0$), and 48.00\% for Monolithic. Relative to Bilateral w/ CC ($\delta=0$), Bilateral w/ CC ($\delta>0$) also improves several precision, stability, and energetic measures, while relative to Bilateral w/o CC it improves success rate, endpoint error, energetic cost, and CCI. Although Bilateral w/o CC shows slightly lower movement variance (0.0080 vs.\ 0.0090, $\Delta=-0.0010$), the broader pattern further strengthens the main conclusion that modular control and delayed inter-module communication remain beneficial beyond the one-active-arm-at-a-time training protocol. For completeness, additional out-of-distribution tests under the original interleaved task setting, using longer reaches and stronger holding perturbations, are reported in Appendix~\ref{app:generalisation}.

\begin{table}[!htbp]
\centering
\caption{Performance under simultaneous bimanual control. Values are mean $\pm$ standard deviation across the matched noise--energy grid and five random seeds. Bilateral w/ CC ($\delta>0$) is the reference condition. Asterisks denote significant differences from this reference after Bonferroni-corrected paired Wilcoxon tests ($p<0.05$). Arrows indicate the direction of improvement, and bold marks the best mean in each row.}
\label{tab:simultaneous_bimanual}
\begin{tabular*}{\textwidth}{@{\extracolsep{\fill}}lcccc@{}}
\toprule
Metric & Bi w/ CC, $\delta>0$ & Bi w/ CC, $\delta=0$ & Bi w/o CC & Mono. \\
\midrule
Success rate $\uparrow$ & \MeanStd{\textbf{83.48\%}}{10.50\%} & \MeanStd{66.15\%$^{*}$}{23.86\%} & \MeanStd{78.41\%$^{*}$}{15.37\%} & \MeanStd{48.00\%$^{*}$}{32.04\%} \\
Endpoint error $\downarrow$ & \MeanStd{\textbf{0.0091}}{0.0019} & \MeanStd{0.0099}{0.0035} & \MeanStd{0.0104$^{*}$}{0.0034} & \MeanStd{0.0101}{0.0029} \\
Trajectory error $\downarrow$ & \MeanStd{\textbf{0.0051}}{0.0006} & \MeanStd{0.0059$^{*}$}{0.0016} & \MeanStd{0.0053}{0.0009} & \MeanStd{0.0060$^{*}$}{0.0012} \\
Movement variance $\downarrow$ & \MeanStd{0.0090}{0.0011} & \MeanStd{0.0091}{0.0018} & \MeanStd{\textbf{0.0080}$^{*}$}{0.0018} & \MeanStd{0.0109$^{*}$}{0.0033} \\
Hold variance $\downarrow$ & \MeanStd{\textbf{1.217e{-}5}}{7.627e{-}6} & \MeanStd{1.922e{-}5$^{*}$}{1.222e{-}5} & \MeanStd{1.244e{-}5}{8.409e{-}6} & \MeanStd{2.908e{-}5$^{*}$}{1.928e{-}5} \\
Energetic cost $\downarrow$ & \MeanStd{\textbf{0.8078}}{1.0075} & \MeanStd{1.1446$^{*}$}{1.0643} & \MeanStd{1.1172$^{*}$}{1.2883} & \MeanStd{1.6468$^{*}$}{1.1513} \\
CCI $\downarrow$ & \MeanStd{\textbf{0.7079}}{0.0303} & \MeanStd{0.7105}{0.0371} & \MeanStd{0.7424$^{*}$}{0.0563} & \MeanStd{0.7591$^{*}$}{0.0510} \\
\bottomrule
\end{tabular*}
\end{table}

\subsection{Learned communication shifts asymmetric contribution patterns}
\label{sec:lesion}
Because the merged output does not impose lateralised or contralateral control, any asymmetric organisation must emerge through learned module-specific motor contributions. We quantify this using a contralaterality index (CI), where positive values indicate stronger contralateral output contributions, and a double-dissociation index (DDI) computed after masking one module's communication and motor-output contribution at inference time. Full definitions are provided in Appendix~\ref{app:metrics}.

Under the matched zero-delay comparison pooled across the shared noise--energy grid, Bilateral w/ CC shows a more contralateral contribution profile than Bilateral w/o CC ($0.0177 \pm 0.0608$ vs.\ $0.0001 \pm 0.0715$), with a significant paired CI difference ($p<0.05$, Appendix~\ref{app:statistics}). Directional lesion summaries show the same qualitative ordering, with a larger average DDI for Bilateral w/ CC ($0.1063 \pm 0.3284$ vs.\ $0.0510 \pm 0.3084$). We therefore interpret these results as a small communication-associated shift in output contributions, with complementary lesion-based evidence remaining statistically inconclusive.

\section{Discussion}
\paragraph{Bilateral organisation as a control architecture.} Our results show that distributing control computation across two recurrent controllers can form a more effective control architecture than a single monolithic controller. Previous work on modular motor control has shown that multiple controllers can organise and combine different control processes across states and control demands, thereby providing more flexible motor-control solutions \citep{wolpertMultiplePairedForward1998,harunoMOSAICModelSensorimotor2001}. Our study extends this understanding to robotic motor control, where the bilateral two-controller architecture showed advantages in task performance, movement stability, and energetic cost under noise and energetic constraints. One possible explanation is that two independent recurrent state spaces provide the system with multiple dynamic representations and computational pathways. The two controllers have their own recurrent parameters and hidden-state dynamics, so even under a shared training objective, their parameters and internal states can evolve differently and produce different dynamic responses. Different control demands therefore do not all need to be handled within a single dynamical system, but can instead be organised and integrated across two recurrent processes, providing the system with richer learned dynamics. This interpretation is consistent with the idea of modular motor-control frameworks, although the current experiments do not directly examine recurrent-state organisation and the underlying mechanism therefore requires further investigation \citep{michaelsGoaldrivenModularNeural2020}. Such an architecture may offer a useful design option for bimanual robotic control problems that require accurate movement and postural stabilisation to be coordinated under multiple motor-control constraints.

\paragraph{Communication changes how control is achieved.} In this work we allowed the two controllers to exchange information through inter-module communication while contributing to bimanual control. In the main experiment, the bilateral controller still maintained a high success rate after communication was removed, but showed higher endpoint error, trajectory error, and movement variance than the communicating controller, indicating reduced movement accuracy and stability. The balance of antagonist activation also changed significantly, suggesting that communication further influences the controller's actuation strategy. For example, under the highest motor-noise condition ($k=0.2$), the communicating and communication-free controllers achieved similar success rates, but the communicating controller retained measurable energetic cost with more selective antagonist recruitment, consistent with continued active control, whereas the communication-free controller shifted towards a lower-energy regime. A related role can also be found in biological motor control, where interhemispheric interactions contribute to the coordination and integration of different motor processes  \citep{serrienDynamicsHemisphericSpecialization2006a,liuzziCoordinationUncoupledBimanual2011}. Allowing communication between controllers may therefore provide a bilateral architecture with a mechanism for adjusting its control strategy, enabling different coordination solutions to emerge under different task and constraint demands. These results indicate that, in a bilateral system that already has strong control capability, communication affects not only how well the task is completed, but also the control strategy and resulting operating regime. 

\paragraph{Communication delay as a design variable.} Further analysis of communication delay showed that the role of communication depends not only on whether it is present, but also on the temporal relationship of information exchange. Compared with zero-delay communication, all tested non-zero delays significantly increased success rate and reduced endpoint error, trajectory error, hold variance, and energetic cost. This is especially noteworthy because communication delay is typically treated as a constraint in conventional and networked control systems, where it can reduce control performance and even affect system stability \citep{richardTimedelaySystemsOverview2003, cloostermanStabilityNetworkedControl2009}, while biological interhemispheric communication is itself also constrained by conduction delay \citep{meyerInhibitoryExcitatoryInterhemispheric1995a,serrienDynamicsHemisphericSpecialization2006a}. One possible explanation is that delayed information changes how the two recurrent controllers interact over time, thereby influencing the dynamics learned by the overall system. Previous work on delayed coupling has shown that delay does more than slow information transmission and can alter the stability, synchronisation, and resulting dynamics of coupled dynamical systems \citep{earlSynchronizationOscillatorNetworks2003,yeungTimeDelayKuramoto1999,reddyTimeDelayInduced1998}. In our model, non-zero delay may reduce excessive dependence between the two controllers on each other's real-time states, allowing them to exchange information while retaining some degree of independent dynamics. It may also alter when information enters the recurrent dynamics, leading optimisation to converge to different control solutions. These results suggest that delays introduced by inter-module communication should not be treated as an inherent performance penalty in modular robotic control. For robotic systems that require multiple interacting controllers, communication delay can be treated as a design variable worth evaluating and adjusting within the architecture.

\paragraph{Extension to simultaneous bimanual control.} Combining reaching with simultaneous stabilisation is also similar to common asymmetric bimanual actions, such as holding a smartphone with one hand while using the other to reach to different locations on the screen. Coordinating both arms is also an active area of research in robotic manipulation \citep{grannenStabilizeAct2023,fuMobileALOHA2024,liuRDT1B2025}. In the original experiment, only one arm performed reaching or holding in each trial, raising the question of whether the advantage of the two-controller architecture mainly arose from this relatively simplified one-active-arm setting. The simultaneous bimanual results showed that when reaching and holding had to be performed concurrently by the two arms, the advantage of the bilateral controller was retained, with the communicating bilateral controller showing advantages over the other controllers in success rate, energetic cost, and several other control metrics.

\paragraph{Emergent asymmetry in output organisation.} Communication also influenced the output organisation of the two controllers. In both bilateral architectures, the architecture itself did not impose fixed control routing, and both controllers could contribute to all 12 muscle commands. Any left–right asymmetry therefore had to emerge during training. After learned communication was introduced, the contralaterality index (CI), which measures the degree of contralateral control, showed a significant but small increase, indicating that the output contributions of the two controllers shifted towards a more contralateral organisation. This direction has a conceptual connection with observations of contralateral control and hemispheric asymmetry in biological motor lateralisation 
\citep{muthaEffectsBrainLateralization2012a,sainburgConvergentModelsHandedness2014a}, and is also consistent with the theoretical view that constrained communication may promote lateralised organisation 
\citep{seoaneOptimalityPressuresLateralization2023a}. Notably, this asymmetry emerged between two controllers with the same architecture and the same sensory access, showing that lateralised output organisation can arise without predefined controller asymmetry. A related perspective exists in human motor development, where hand dominance leads to a preference for using one hand over the other for certain tasks, resulting in asymmetric practice \citep{aracArmDominanceEmergent2026}. Such asymmetric experience may in turn reinforce hemispheric asymmetry, providing a possible biological parallel to the stronger lateralised organisation observed with communication in our model. Communication may therefore influence not only control performance and actuation strategy, but also how control contributions are organised across modules. The magnitude of the CI change was small, and although the lesion-based analysis showed the same directional trend, the difference between architectures was not significant. Therefore, the evidence supports an emerging asymmetric output-contribution pattern, but not stable functional specialisation, which would require clearer controller-level asymmetry with measurable functional consequences.

\paragraph{Interpreting control under high motor noise.} The high-noise regime revealed a phenomenon worth discussing. In real robotic systems, hardware degradation and sensing uncertainty can increase control uncertainty and movement variability, making motor behaviour less stable and less predictable \citep{kumaranRemainingUsefulLife2023,xiaoRobustTrackingControl2020}. In the signal-dependent motor-noise model used here, noise magnitude increases with motor command, so reducing motor commands also reduces the noise injected into the system \citep{harrisSignaldependentNoiseDetermines1998a}. Our results suggest that this property can encourage a low-output strategy, in which the controller reduces activation enough to satisfy the original holding success criterion without necessarily developing stronger active stabilisation. The apparent rebound in success of the monolithic controller at the highest noise level illustrates this behaviour. When the holding perturbation was increased, this rebound was substantially reduced, indicating that the previously high success could not be maintained under stronger external disturbance and was consistent with an explanation based on a low-output strategy rather than stronger stabilising control. This observation is also consistent with optimal feedback control and sensorimotor control theories, which view movement performance as the outcome of interacting accuracy, effort, and uncertainty-related objectives \citep{todorovOptimalFeedbackControl2002a,harrisSignaldependentNoiseDetermines1998a}.

\paragraph{Limitations and future work}
First, some interpretations of the bilateral architecture, communication, and communication delay are based on behavioural and control-level evidence rather than direct analysis of the underlying recurrent mechanisms. We propose that separate recurrent state spaces may support richer learned dynamics, and that non-zero delay may alter how information is integrated across controllers, but these mechanisms have not been directly tested. Likewise, learned communication increased the contralaterality index, but the effect was small and the lesion-based DDI did not differ significantly between architectures. Future work could use hidden-state and state-space analyses, information-flow measures, and targeted communication or lesion interventions to test whether the controllers develop distinct recurrent dynamics and whether these differences have causal functional consequences. Second, the experiments cover a limited range of communication delays, motor-noise gains, and energetic penalties. Although all tested non-zero delays consistently outperformed zero-delay communication, the current parameter ranges do not establish whether this relationship persists more broadly, whether an optimal delay exists, or whether there are transition points or nonlinear interactions among constraints. Future work could expand the parameter ranges and sampling resolution and examine longer, variable, or asymmetric communication delays. Third, although the simultaneous bimanual experiment and the longer-reaching and stronger-perturbation out-of-distribution tests extend the original interleaved setting, the task family remains centred on reaching and postural holding. More complex object interactions, mechanical coupling between the arms, and dexterous bimanual manipulation were not examined. Future work should test these architectures in cooperative manipulation, mechanically coupled tasks, and other settings that require sustained bimanual coordination. Finally, all experiments were conducted in the MotorNet musculoskeletal simulation environment, so the applicability of the findings to other robotic systems and physical hardware remains to be established. The six-muscle arm model does not reflect most conventional robotic platforms, which typically operate through joint-level control. Real robots also introduce additional factors, including actuator and transmission dynamics as well as sensing uncertainty. Future work should therefore evaluate the bilateral architecture, inter-module communication, and communication delay in other simulation platforms, such as MuJoCo \citep{todorovMuJoCoPhysicsEngine2012}, as well as on physical robotic systems.

\section{Conclusion} 
Inspired by the bilateral organisation of the brain and sensorimotor system, this study examined how bilateral controller architecture, inter-module communication, and communication delay influence bimanual robotic motor control under noise and energetic constraints. We show that a bilateral controller architecture outperforms a parameter-matched monolithic controller under noise and energetic constraints. Inter-module communication further improves movement accuracy and stability, alters actuation strategy and output organisation, and its timing also matters, with all tested non-zero delays outperforming zero-delay communication. Together, our findings suggest that, under noise and energetic pressure, bilateral two-controller organisation provides a useful architectural choice for bimanual robotic control. Beyond the architecture itself, effective control also depends on how the modules communicate and coordinate over time.

\newpage
\bibliographystyle{tmlr}
\bibliography{lrbrain_related_work_shortlist}

\newpage
\appendix

\section{Controller architecture}
\subsection{Bilateral controller architecture}
\label{app:architecture}
Let $x_t\in\mathbb{R}^{48}$ denote the shared sensory observation, comprising proprioceptive and task-related information for both arms. Bilateral w/ CC contains two GRU modules with hidden size 64. Each module receives the shared sensory observation together with a learned projection of the other module's communicated hidden state. The recurrent updates are
\begin{align}
h_t^L &= \mathrm{GRU}_L\left([x_t,C_{R\to L}\bar h_t^R],h_{t-1}^L\right),\\
h_t^R &= \mathrm{GRU}_R\left([x_t,C_{L\to R}\bar h_t^L],h_{t-1}^R\right),
\end{align}
where $C_{R\to L}$ and $C_{L\to R}$ are learned $64\times64$ inter-module projection matrices. For non-zero communication delay $\delta$, the communicated state is drawn from the corresponding buffered past hidden state. The simulation timestep is 10 ms, so $\delta\in\{5,10,20\}$ corresponds to 50, 100, and 200 ms, respectively.

The two module-specific readouts contribute to the same 12-dimensional motor-command vector:
\begin{equation}
u_t=\sigma\left(W_Lh_t^L+W_Rh_t^R+b\right),
\end{equation}
where $W_L,W_R\in\mathbb{R}^{12\times64}$ are learned readout matrices, $b\in\mathbb{R}^{12}$ is a shared bias, and $\sigma$ constrains each muscle command to $[0,1]$. The first six components of $u_t$ actuate the left arm and the remaining six actuate the right arm. No fixed routing or lateralised output mapping is imposed and both modules can contribute to all 12 muscle commands.

Signal-dependent motor noise is applied after the deterministic readout:
\begin{equation}
\widetilde u_t=
\mathrm{clip}_{[0,1]}
\left(u_t\odot(1+k\epsilon_t)\right),
\qquad
\epsilon_t\sim\mathcal N(0,I_{12}),
\end{equation}
where $k$ controls noise magnitude. Bilateral w/o CC retains the same two GRU-64 modules and merged readout but disables $C_{R\to L}$ and $C_{L\to R}$. The Monolithic controller instead uses a single GRU-128 receiving the same sensory observation and mapping directly to the same 12-dimensional muscle-command space.

\subsection{Parameter-matching controls}
\label{app:param_match}
The primary Bilateral w/o CC communication ablation removes the cross-module projections and therefore has fewer parameters than Bilateral w/ CC. To test whether the main results depend on this parameter-count difference, we evaluate two additional parameter-matching controller architectures as a further control analysis. Bilateral w/o CC remains the primary communication ablation, while these additional architectures assess the robustness of the results to parameter matching.

\begin{table}[h]
\centering
\caption{Trainable-parameter breakdown for the three primary controller architectures and two parameter-matching control architectures. SelfProj exactly matches Bilateral w/ CC in parameter count while removing cross-module information exchange. Monolithic-137 is the nearest larger integer-width monolithic match.}
\label{tab:param_counts}
\begin{tabular*}{\textwidth}{@{\extracolsep{\fill}}lrrrrr@{}}
\toprule
Parameter group & Bi w/ CC & Bi w/o CC & SelfProj & Monolithic-128 & Monolithic-137 \\
\midrule
GRU input weights     & 43,008 & 43,008 & 43,008 & 18,432 & 19,728 \\
GRU recurrent weights & 24,576 & 24,576 & 24,576 & 49,152 & 56,307 \\
GRU biases            & 768    & 768    & 768    & 768    & 822 \\
Projection matrices   & 8,192  & 0      & 8,192  & 0      & 0 \\
Readout and bias      & 1,548  & 1,548  & 1,548  & 1,548  & 1,656 \\
\midrule
\textbf{Total}        & \textbf{78,092} & \textbf{69,900} & \textbf{78,092} & \textbf{69,900} & \textbf{78,513} \\
\bottomrule
\end{tabular*}
\end{table}

\paragraph{Controller architectures.}
The \textbf{SelfProj} controller architecture replaces each cross-module projection with an equally sized projection of the same module's previous hidden state. It therefore retains the same auxiliary projection pathway and exactly matches Bilateral w/ CC at 78,092 trainable parameters, while preventing information exchange between modules. The \textbf{Monolithic-137} controller architecture increases the monolithic GRU hidden size from 128 to 137, yielding 78,513 parameters, 421 (0.54\%) more than Bilateral w/ CC. Because GRU hidden size is integer-valued, an exact monolithic parameter match is not available. GRU-137 is the nearest larger-width match. Table~\ref{tab:param_counts} gives the full parameter breakdown.

\paragraph{Evaluation protocol.}
Both parameter-matching controls use the same four energetic-cost weights, four motor-noise levels, ten random seeds, and interleaved reaching and holding protocol as the main experiment. Comparisons therefore use $N=160$ matched $(\lambda,k,\mathrm{seed})$ combinations. As in Table~\ref{tab:main_results}, the $\delta>0$ Bilateral w/ CC reference is first averaged over $\delta\in\{5,10,20\}$ within each matched $(\lambda,k,\mathrm{seed})$ setting before paired comparison.

\begin{table}[h]
\centering
\caption{Parameter-matching results over $N=160$ matched $(\lambda,k,\mathrm{seed})$ combinations. Bilateral w/ CC ($\delta>0$) is the reference condition. The $\delta=0$ column provides context for the main delay comparison, while SelfProj and Monolithic-137 are the parameter-matching controls. Asterisks denote a significant difference from the reference under Bonferroni-corrected paired Wilcoxon tests ($p<0.05$). Bold indicates the best mean in each row. Exact corrected values are reported in Appendix~\ref{app:statistics}.}
\label{tab:param_match_results}
\begin{tabular*}{\textwidth}{@{\extracolsep{\fill}}lcccc@{}}
\toprule
Metric & Bilateral w/ CC, $\delta>0$ & Bilateral w/ CC, $\delta=0$ & SelfProj & Monolithic-137 \\
\midrule
Success rate $\uparrow$ & \MeanStd{\textbf{88.81\%}}{10.01\%} & \MeanStd{77.53\%$^{*}$}{21.99\%} & \MeanStd{78.13\%$^{*}$}{20.96\%} & \MeanStd{63.99\%$^{*}$}{30.32\%} \\
Endpoint error $\downarrow$ & \MeanStd{\textbf{0.0085}}{0.0017} & \MeanStd{0.0091$^{*}$}{0.0022} & \MeanStd{0.0089}{0.0026} & \MeanStd{0.0087}{0.0018} \\
Trajectory error $\downarrow$ & \MeanStd{\textbf{0.0022}}{0.0003} & \MeanStd{0.0025$^{*}$}{0.0005} & \MeanStd{0.0025$^{*}$}{0.0005} & \MeanStd{0.0026$^{*}$}{0.0004} \\
Movement variance $\downarrow$ & \MeanStd{\textbf{0.0081}}{0.0013} & \MeanStd{0.0092$^{*}$}{0.0038} & \MeanStd{0.0091$^{*}$}{0.0022} & \MeanStd{0.0104$^{*}$}{0.0031} \\
Hold variance $\downarrow$ & \MeanStd{\textbf{1.054e{-}5}}{6.723e{-}6} & \MeanStd{1.556e{-}5$^{*}$}{1.471e{-}5} & \MeanStd{1.492e{-}5$^{*}$}{1.017e{-}5} & \MeanStd{2.154e{-}5$^{*}$}{1.449e{-}5} \\
Energetic cost $\downarrow$ & \MeanStd{\textbf{1.0875}}{1.6950} & \MeanStd{1.3196$^{*}$}{1.7233} & \MeanStd{1.8839$^{*}$}{2.1055} & \MeanStd{2.6096$^{*}$}{2.2634} \\
CCI $\downarrow$ & \MeanStd{\textbf{0.7175}}{0.0369} & \MeanStd{0.7340$^{*}$}{0.0401} & \MeanStd{0.7261$^{*}$}{0.0418} & \MeanStd{0.7592$^{*}$}{0.0440} \\
\bottomrule
\end{tabular*}
\end{table}

As shown in Table~\ref{tab:param_match_results}, both parameter-matching controls support the main conclusions. SelfProj differs significantly from Bilateral w/ CC ($\delta>0$) on six of seven metrics, with endpoint error as the only non-significant comparison, showing that exact parameter matching and an equally sized local projection pathway do not reproduce the delayed communication result. Monolithic-137 shows the same six-of-seven pattern despite having slightly more parameters than Bilateral w/ CC, indicating that the modularity advantage is not explained by the parameter-count difference in the original Monolithic-128 comparison.

\section{Evaluation metrics}
\label{app:metrics}
All evaluation metrics are computed on held-out inference rollouts under the same task geometry used during training, unless otherwise stated. For each trained model, metrics are averaged across trials, then summarised across matched experimental conditions in the noise--energy grid and random seeds.

\paragraph{Task success.}
Reach success indicates that the final endpoint error after a reaching trial is below the reach tolerance. Hold success indicates that the hold-phase endpoint variance is below the stability tolerance. In the main experiments, the reach threshold is $0.02\,\mathrm{m}$ and the hold-variance threshold is $2.5\times 10^{-5}\,\mathrm{m}^2$. Because reach and hold are evaluated in separate but matched inference rollouts, the reported success rate is computed from the corresponding success probabilities. Let $p_{\mathrm{reach}}^{L}$ and $p_{\mathrm{reach}}^{R}$ denote the success rates for left-active and right-active reach trials, and let $p_{\mathrm{hold}}^{L}$ and $p_{\mathrm{hold}}^{R}$ denote the hold-success rates for the left and right arms. The reported success rate is thus:
\begin{equation}
\mathrm{SR} = \frac{1}{2}\left(p_{\mathrm{reach}}^{L}p_{\mathrm{hold}}^{R} + p_{\mathrm{reach}}^{R}p_{\mathrm{hold}}^{L}\right).
\end{equation}
This factorised definition gives a fairer controller-level comparison because it avoids introducing an arbitrary one-to-one pairing between independent reach and hold batches while still requiring success in both task components for each matched experimental condition and seed.

\paragraph{Kinematic and stability metrics.}
Endpoint error is the Euclidean distance between the final hand position and the target at the end of the reach phase. Trajectory error is the mean absolute perpendicular deviation of the reach trajectory from the straight line between start and target. Movement variance is the scaled velocity variance during the movement phase, computed as the time-variance of the planar hand velocity summed across Cartesian dimensions, averaged across trials, and scaled by a constant factor of 0.1 in the reported experiments. Hold variance is the endpoint-position variance during the holding phase,
\begin{equation}
\mathrm{Var}_{\mathrm{hold}} =
\frac{1}{B}\sum_{b=1}^{B}\left[\mathrm{Var}_{t}(x_{t,b})+\mathrm{Var}_{t}(y_{t,b})\right],
\end{equation}
where $B$ is the number of trials in the batch. Lower values indicate better precision or stability for all four metrics.

\paragraph{Actuation metrics.}
The reported energetic cost is the mean squared muscle activation summed over muscles and averaged over time and trials:
\begin{equation}
E_{\mathrm{eval}} =
\frac{1}{BT}\sum_{b=1}^{B}\sum_{t=1}^{T}\sum_{i=1}^{M} u_{t,b,i}^{2}.
\end{equation}
The co-contraction index (CCI) measures simultaneous activation of antagonist muscle pairs. For an antagonist pair with activations $a_1$ and $a_2$,
\begin{equation}
\mathrm{CCI}_{\mathrm{pair}} =
\frac{2\min(a_1,a_2)}{a_1+a_2+\epsilon},
\end{equation}

and the reported CCI averages this quantity over antagonist pairs, time, and trials. This formulation follows the ratio-based Falconer--Winter co-contraction index and measures the relative balance of activation within each antagonist pair \citep{falkIncreasedCocontractionReaction2022,liHowWellCommonly2021}. CCI approaches one when the two antagonist activations are similar and approaches zero when one muscle dominates the pair. Because the index is normalised by total pair activation, it does not by itself quantify absolute muscle activation or joint stiffness: two weakly activated antagonist muscles can still produce a high CCI if their activations are similar. We therefore interpret CCI together with energetic cost, with CCI describing relative antagonist co-activation and energetic cost reflecting the overall magnitude of muscle activation.

\paragraph{Asymmetry and lesion metrics.}
For each module $m\in\{L,R\}$, the contralaterality index (CI) compares the mean absolute contralateral versus ipsilateral output contributions measured during inference rollouts:
\begin{equation}
\mathrm{CI}_m = \frac{C_m-I_m}{C_m+I_m+\epsilon},
\qquad
\mathrm{CI}=\frac{1}{2}(\mathrm{CI}_L+\mathrm{CI}_R),
\end{equation}
where positive values indicate stronger contralateral output contributions. 

Directional lesion effects are measured with inference-time module lesions. For a left-module lesion, the left module's outgoing inter-module communication and motor-output contribution are set to zero. The right-module lesion is defined analogously. Trained-model parameters are not updated after lesioning, and the intact, left-module-lesioned, and right-module-lesioned conditions are evaluated on the same reach and hold protocol. Lesion effects are summarised by a double-dissociation index (DDI). For the main DDI, the arm error used in each lesion condition is a normalised reach--hold combination,

\begin{equation}
e_{\mathrm{arm}} = \frac{1}{2}\left(\frac{e_{\mathrm{final}}}{0.02} + \frac{e_{\mathrm{hold}}}{0.005}\right),
\end{equation}
where $e_{\mathrm{final}}$ is final reach endpoint error and $e_{\mathrm{hold}}$ is mean hold-phase positional error. Let $\Delta^{L}_{\mathrm{contra}}$ and $\Delta^{L}_{\mathrm{ipsi}}$ be the increases in contralateral and ipsilateral arm error after lesioning the left module, and define $\Delta^{R}_{\mathrm{contra}}$ and $\Delta^{R}_{\mathrm{ipsi}}$ analogously for right-module lesions. We compute
\begin{equation}
\mathrm{DDI}_{L}=
\frac{\Delta^{L}_{\mathrm{contra}}-\Delta^{L}_{\mathrm{ipsi}}}
{|\Delta^{L}_{\mathrm{contra}}|+|\Delta^{L}_{\mathrm{ipsi}}|+\epsilon},
\qquad
\mathrm{DDI}_{R}=
\frac{\Delta^{R}_{\mathrm{contra}}-\Delta^{R}_{\mathrm{ipsi}}}
{|\Delta^{R}_{\mathrm{contra}}|+|\Delta^{R}_{\mathrm{ipsi}}|+\epsilon},
\end{equation}
and report $\mathrm{DDI}=(\mathrm{DDI}_{L}+\mathrm{DDI}_{R})/2$. Positive DDI indicates that lesioning a module impairs the contralateral arm more than the ipsilateral arm.

\section{Delay-specific analysis}
\label{app:delay}
To test whether the aggregate $\delta>0$ result is driven by a single favourable delay, we analyse $\delta=5$, 10, and 20 separately.

\begin{table}[h]
\centering
\caption{Delay-specific results for Bilateral w/ CC (mean $\pm$ standard deviation, $N=160$ matched $(\lambda,k,\mathrm{seed})$ combinations per delay). Each positive-delay condition is compared with the matched $\delta=0$ condition using a two-sided paired Wilcoxon signed-rank test. Asterisks denote significance after Bonferroni correction across the three delay contrasts within each metric ($p<0.05$).}
\label{tab:delay_specific}
\begin{tabular*}{\textwidth}{@{\extracolsep{\fill}}lcccc@{}}
\toprule
Metric & $\delta=0$ & $\delta=5$ & $\delta=10$ & $\delta=20$ \\
\midrule
Success rate $\uparrow$ & \MeanStd{77.53\%}{21.99\%} & \MeanStd{87.99\%$^{*}$}{14.93\%} & \MeanStd{88.50\%$^{*}$}{11.21\%} & \MeanStd{89.95\%$^{*}$}{14.09\%} \\
Endpoint error $\downarrow$ & \MeanStd{0.00909}{0.00215} & \MeanStd{0.00817$^{*}$}{0.00179} & \MeanStd{0.00862$^{*}$}{0.00252} & \MeanStd{0.00868$^{*}$}{0.00213} \\
Trajectory error $\downarrow$ & \MeanStd{0.00250}{0.00049} & \MeanStd{0.00230$^{*}$}{0.00046} & \MeanStd{0.00222$^{*}$}{0.00041} & \MeanStd{0.00211$^{*}$}{0.00042} \\
Hold variance $\downarrow$ & \MeanStd{1.5564e{-}5}{1.4706e{-}5} & \MeanStd{1.1748e{-}5$^{*}$}{9.3180e{-}6} & \MeanStd{1.0447e{-}5$^{*}$}{6.6546e{-}6} & \MeanStd{9.4241e{-}6$^{*}$}{7.4621e{-}6} \\
Energetic cost $\downarrow$ & \MeanStd{1.3196}{1.7233} & \MeanStd{1.1306$^{*}$}{1.6569} & \MeanStd{1.0449$^{*}$}{1.7008} & \MeanStd{1.0871$^{*}$}{1.7580} \\
\bottomrule
\end{tabular*}
\end{table}
Each delay condition contains the same $N=160$ matched $(\lambda,k,\mathrm{seed})$ combinations as $\delta=0$. We pair each positive-delay observation with its $\delta=0$ counterpart and use two-sided paired Wilcoxon signed-rank tests, with Bonferroni correction across the three delay contrasts within each metric.

As shown in Table~\ref{tab:delay_specific}, all 15 positive-delay contrasts remain significant after correction, with the largest corrected $p$-value equal to 0.0353. Each tested non-zero delay increases success rate and reduces endpoint error, trajectory error, hold variance, and energetic cost relative to zero-delay communication. The aggregate $\delta>0$ result is therefore not driven by a single delay setting but is consistent across the tested range. These results do not identify an optimal delay or a specific mechanism, but they consistently show improved learned control across all tested non-zero delays.

\section{Generalisation to unseen task conditions}
\label{app:generalisation}
We additionally test whether the trained controllers generalise to stronger task demands without retraining. We test two out-of-distribution conditions: \textit{reach-16}, which increases the reach radius to 0.16 m and the number of targets from eight to twelve, and \textit{hold-2N}, which increases the per-axis standard deviation of the Gaussian holding perturbation to 2 N. 

\begin{figure}[h]
\centering
\includegraphics[width=0.8\linewidth]{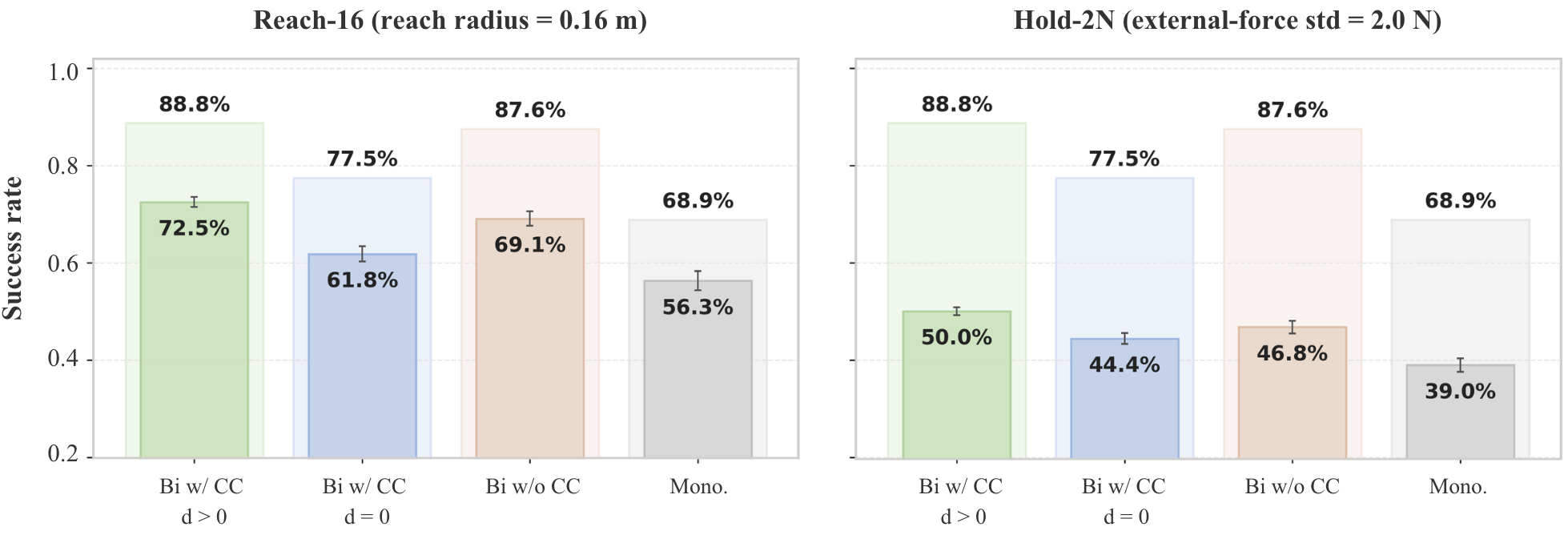}
\caption{Generalisation results for success rate on unseen reach and hold conditions. Filled bars show out-of-distribution performance, and faint bars show the matched in-distribution baseline. Bar labels report mean success rates, and error bars on the out-of-distribution bars show the standard error of the mean over $N=160$ matched $(\lambda,k,\mathrm{seed})$ values. For Bilateral w/ CC, zero-delay trained models ($\delta=0$) are separated from delayed trained models ($\delta>0$), delayed values are first averaged over $\delta\in\{5,10,20\}$ within each matched $(\lambda,k,\mathrm{seed})$ setting. On \textit{reach-16}, success uses a fixed reach threshold of 20 mm. On \textit{hold-2N}, the Gaussian holding perturbation has per-axis standard deviation $2\,\mathrm{N}$.}
\label{fig:generalisation_unseen}
\end{figure}

Figure~\ref{fig:generalisation_unseen} shows that Bilateral w/ CC ($\delta>0$) achieves the highest success rate under both out-of-distribution conditions, followed by Bilateral w/o CC, Bilateral w/ CC ($\delta=0$), and Monolithic. These results complement the simultaneous bimanual experiment in Section~\ref{sec:bimanual}: the latter tests whether the main findings extend to concurrent bilateral control, whereas these analyses test robustness to increased movement distance and perturbation strength within the original task structure.

\section{Statistical tests and detailed results}
\label{app:statistics}

Unless stated otherwise, all comparisons use $N=160$ matched $(\lambda,k,\mathrm{seed})$ combinations and two-sided paired Wilcoxon signed-rank tests. For the $\delta>0$ reference condition, values are first averaged over $\delta\in\{5,10,20\}$ within each matched $(\lambda,k,\mathrm{seed})$ setting before paired testing. For both the main comparison in Table~\ref{tab:main_results} and the parameter-matching comparison in Table~\ref{tab:param_match_results}, the three comparator $p$-values within each metric are Bonferroni corrected. Table~\ref{tab:corrected_pvalues} reports the corrected values. ``n.s.'' denotes $p\geq0.05$.

\begin{table}[!htbp]
\centering
\caption{Bonferroni-corrected paired Wilcoxon $p$-values relative to Bilateral w/ CC ($\delta>0$). The first three comparator columns correspond to the three-comparator family used in Table~\ref{tab:main_results}. SelfProj and Monolithic-137 belong to the separate three-comparator family used in Table~\ref{tab:param_match_results}, together with Bilateral w/ CC ($\delta=0$).}
\label{tab:corrected_pvalues}
\begin{tabular*}{\textwidth}{@{\extracolsep{\fill}}lccccc@{}}
\toprule
Metric & Bi w/ CC, $\delta=0$ & Bi w/o CC & Monolithic-128 & SelfProj & Monolithic-137 \\
\midrule
Success rate      & $2.95{\times}10^{-10}$ & $1.000$ (n.s.) & $1.06{\times}10^{-13}$ & $6.86{\times}10^{-9}$ & $1.80{\times}10^{-15}$ \\
Endpoint error    & $8.60{\times}10^{-4}$  & $1.22{\times}10^{-19}$ & $7.28{\times}10^{-3}$ & $1.000$ (n.s.) & $1.000$ (n.s.) \\
Trajectory error  & $3.63{\times}10^{-10}$ & $3.83{\times}10^{-18}$ & $1.48{\times}10^{-12}$ & $2.35{\times}10^{-8}$ & $2.55{\times}10^{-14}$ \\
Movement variance & $6.99{\times}10^{-6}$  & $1.31{\times}10^{-7}$  & $1.82{\times}10^{-10}$ & $7.14{\times}10^{-9}$ & $1.03{\times}10^{-13}$ \\
Hold variance     & $2.67{\times}10^{-9}$  & $0.942$ (n.s.) & $4.23{\times}10^{-14}$ & $9.45{\times}10^{-9}$ & $1.15{\times}10^{-16}$ \\
Energetic cost    & $2.11{\times}10^{-7}$  & $0.512$ (n.s.) & $7.79{\times}10^{-13}$ & $1.78{\times}10^{-21}$ & $9.36{\times}10^{-21}$ \\
CCI               & $2.10{\times}10^{-8}$  & $3.72{\times}10^{-12}$ & $2.70{\times}10^{-16}$ & $4.57{\times}10^{-3}$ & $4.39{\times}10^{-15}$ \\
\bottomrule
\end{tabular*}
\end{table}

\paragraph{Simultaneous bimanual control.}
The simultaneous bimanual comparison in Table~\ref{tab:simultaneous_bimanual} is the exception to the default $N=160$ setting above. It uses $N=80$ matched $(\lambda,k,\mathrm{seed})$ combinations from four energy penalties, four noise levels, and five random seeds. As above, the $\delta>0$ reference is formed by averaging $\delta\in\{5,10,20\}$ within each matched combination before testing. Each comparator is then paired with this reference over 80 combinations using a two-sided Wilcoxon signed-rank test, and the three comparator $p$-values are Bonferroni corrected separately within each metric. Table~\ref{tab:simultaneous_corrected_pvalues} reports the corrected values. Movement variance and CCI, which were not retained in the original summary JSON files, were recomputed from the saved checkpoints using a fixed-seed held-out replay. The other metrics use the saved evaluation results.

\begin{table}[!htbp]
\centering
\caption{Bonferroni-corrected paired Wilcoxon $p$-values for the simultaneous bimanual comparison in Table~\ref{tab:simultaneous_bimanual}, relative to Bilateral w/ CC ($\delta>0$). Each row forms a three-comparator correction family.}
\label{tab:simultaneous_corrected_pvalues}
\begin{tabular*}{\textwidth}{@{\extracolsep{\fill}}lccc@{}}
\toprule
Metric & Bi w/ CC, $\delta=0$ & Bi w/o CC & Monolithic-128 \\
\midrule
Success rate      & $2.98{\times}10^{-9}$  & $8.90{\times}10^{-3}$ & $5.52{\times}10^{-13}$ \\
Endpoint error    & $0.664$ (n.s.)          & $3.07{\times}10^{-4}$ & $0.0938$ (n.s.) \\
Trajectory error  & $2.84{\times}10^{-4}$  & $0.543$ (n.s.)        & $5.46{\times}10^{-6}$ \\
Movement variance & $1.000$ (n.s.)          & $3.42{\times}10^{-5}$ & $1.21{\times}10^{-4}$ \\
Hold variance     & $2.84{\times}10^{-10}$ & $1.000$ (n.s.)        & $1.22{\times}10^{-13}$ \\
Energetic cost    & $2.97{\times}10^{-12}$ & $7.76{\times}10^{-7}$ & $1.37{\times}10^{-12}$ \\
CCI               & $1.000$ (n.s.)          & $1.03{\times}10^{-6}$ & $3.90{\times}10^{-9}$ \\
\bottomrule
\end{tabular*}
\end{table}

\FloatBarrier
\paragraph{Direct matched-parameter modularity contrast.}
To test the modularity claim directly, we compare Bilateral w/o CC with Monolithic-128 over $N=160$ matched $(\lambda,k,\mathrm{seed})$ combinations. These architectures each contain exactly 69,900 trainable parameters (Table~\ref{tab:param_counts}), so this contrast isolates the effect of splitting the recurrent controller into two modules without a parameter-count difference. Table~\ref{tab:nocc_vs_mono} reports two-sided paired Wilcoxon tests. We report raw $p$-values and Bonferroni-adjusted values across seven displayed metrics. Differences are computed as Bilateral w/o CC minus Monolithic-128.

\begin{table}[!htbp]
\centering
\caption{Direct matched-parameter comparison between Bilateral w/o CC and Monolithic-128. Both architectures contain 69,900 trainable parameters and are compared over $N=160$ matched $(\lambda,k,\mathrm{seed})$ combinations using two-sided paired Wilcoxon signed-rank tests. $\Delta$ is Bilateral w/o CC minus Monolithic-128 and is reported in percentage points for success rate. $p_{\mathrm{Bonf.}}$ applies Bonferroni correction across seven metrics.}
\label{tab:nocc_vs_mono}
\begin{tabular*}{\textwidth}{@{\extracolsep{\fill}}lccccc@{}}
\toprule
Metric & Bi w/o CC & Monolithic-128 & $\Delta$ & Raw $p$ & $p_{\mathrm{Bonf.}}$ \\
\midrule
Success rate      & \MeanStd{87.59\%}{13.82\%} & \MeanStd{68.91\%}{27.81\%} & $+18.69$ pp & $6.57{\times}10^{-10}$ & $4.60{\times}10^{-9}$ \\
Endpoint error    & \MeanStd{0.01166}{0.00406} & \MeanStd{0.00912}{0.00227} & $+0.00254$ & $6.23{\times}10^{-12}$ & $4.36{\times}10^{-11}$ \\
Trajectory error  & \MeanStd{0.00456}{0.00250} & \MeanStd{0.00258}{0.00048} & $+0.00198$ & $4.36{\times}10^{-13}$ & $3.05{\times}10^{-12}$ \\
Movement variance & \MeanStd{0.00934}{0.00265} & \MeanStd{0.00987}{0.00304} & $-0.00053$ & $7.07{\times}10^{-3}$ & $4.95{\times}10^{-2}$ \\
Hold variance     & \MeanStd{1.128e{-}5}{8.132e{-}6} & \MeanStd{1.867e{-}5}{1.338e{-}5} & $-7.390{\times}10^{-6}$ & $1.32{\times}10^{-14}$ & $9.24{\times}10^{-14}$ \\
Energetic cost    & \MeanStd{1.1930}{1.8288} & \MeanStd{1.7666}{1.9085} & $-0.5736$ & $5.25{\times}10^{-18}$ & $3.68{\times}10^{-17}$ \\
CCI               & \MeanStd{0.7473}{0.0340} & \MeanStd{0.7606}{0.0466} & $-0.01334$ & $2.56{\times}10^{-4}$ & $1.79{\times}10^{-3}$ \\
\bottomrule
\end{tabular*}
\end{table}

\FloatBarrier
All seven differences remain significant after correction. The communication-free modular controller has higher success rate and lower movement variance, hold variance, energetic cost, and CCI than Monolithic-128, but higher endpoint and trajectory errors. Thus, this equal-parameter comparison supports modularity as the dominant source of the performance gain, while showing that the two architectures occupy different precision--stability--effort trade-offs, without either uniformly dominating every metric.

\paragraph{Delay-specific contrasts.}
For Table~\ref{tab:delay_specific}, each positive-delay condition is paired with its matched $\delta=0$ counterpart. The three delay contrasts ($\delta=5$, 10, and 20 versus $\delta=0$) are Bonferroni corrected within each metric. All 15 remain significant after correction, with the largest corrected value $p=0.0353$.

\paragraph{Asymmetry relative to zero.}
For each delay condition, CI and DDI are tested against zero over the 160 matched $(\lambda,k,\mathrm{seed})$ combinations using two-sided one-sample Wilcoxon signed-rank tests, with Bonferroni correction across the eight delay-by-index tests. Both indices differ significantly from zero at every delay condition. Corrected $p$-values range from $8.44{\times}10^{-5}$ to $0.01484$ for CI and from $1.46{\times}10^{-5}$ to $0.00496$ for DDI, with corresponding rank-biserial effect sizes of 0.284--0.402 and 0.312--0.435, respectively.

\paragraph{Communication-associated asymmetry at zero delay.}
We compare Bilateral w/ CC at $\delta=0$ with Bilateral w/o CC over the same 160 matched $(\lambda,k,\mathrm{seed})$ combinations. CI quantifies asymmetry in learned output contributions, whereas DDI provides a lesion-based measure of directional functional asymmetry. We use two-sided paired Wilcoxon signed-rank tests, paired bootstrap 95\% confidence intervals for the mean difference, and paired rank-biserial effect sizes. Inter-module communication produces a significant increase in CI, whereas the corresponding difference in DDI is not significant (Table~\ref{tab:asymmetry_stats}).

\begin{table}[h]
\centering
\caption{Matched asymmetry comparison between Bilateral w/ CC at $\delta=0$ and Bilateral w/o CC. Differences are computed as CC minus no-CC. Bootstrap intervals are 95\% confidence intervals for the paired mean difference, and $r_{\mathrm{rb}}$ denotes the paired rank-biserial effect size.}
\label{tab:asymmetry_stats}
\begin{tabular*}{\textwidth}{@{\extracolsep{\fill}}lrrrrrr@{}}
\toprule
Metric & CC mean & No-CC mean & Difference & 95\% CI & $p$ & $r_{\mathrm{rb}}$ \\
\midrule
CI  & 0.0177 & 0.0001 & 0.0176 & [0.0025, 0.0323] & 0.0293 & 0.199 \\
DDI & 0.1063 & 0.0510 & 0.0553 & [$-0.0223$, 0.1305] & 0.1048 & 0.148 \\
\bottomrule
\end{tabular*}
\end{table}

These results support an asymmetric output-contribution pattern associated with inter-module communication, but do not provide evidence for a corresponding difference between controller architectures in directional lesion asymmetry.

\section{High-noise diagnostic}
\label{app:high_noise}

The high-noise result for trained models using the monolithic controller architecture requires caution. In the main experiment, these trained models show an apparent rebound in success rate at the highest motor-noise condition ($k=0.2$). Because motor noise is signal-dependent, this rebound need not indicate stronger active control. If a trained model learns to issue very small muscle commands, it also reduces both the evaluated energetic cost and the amount of injected motor noise. Under the original holding condition, where the external force is sampled from a zero-mean Gaussian distribution with per-axis standard deviation $0.5\,\mathrm{N}$, such a low-output strategy can still satisfy the hold-variance threshold. Thus, success rate can remain high even without robust active stabilisation.

\begin{figure}[h]
\centering
\includegraphics[width=0.60\linewidth]{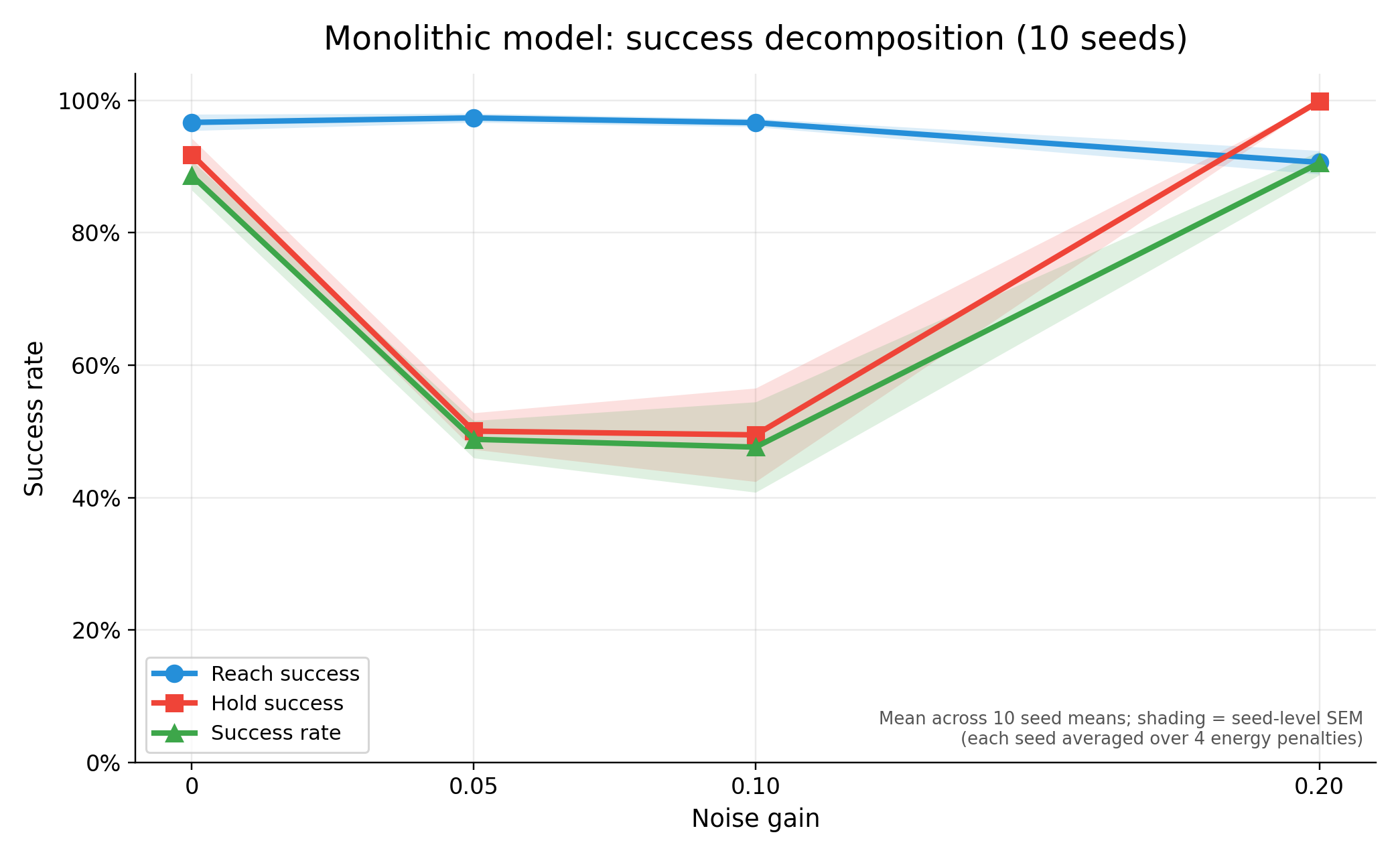}
\caption{Success decomposition for the monolithic baseline under the original Gaussian hold perturbation with per-axis standard deviation $0.5\,\mathrm{N}$. Lines show mean reach success, hold success, and success rate across motor-noise levels, averaged over ten seed-level means, with each seed first averaged over the four energy-penalty settings. Shaded regions indicate $\pm$ one standard error of the mean across seeds.}
\label{fig:app_high_noise_success_decomp}
\end{figure}
    
Figure~\ref{fig:app_high_noise_success_decomp} decomposes the monolithic result into reach and hold success. Reach success remains high across noise levels, whereas hold success drops at intermediate noise and then recovers at $k=0.2$. Thus, the high-noise rebound in success rate is driven mainly by the hold component, despite no corresponding improvement in reaching.

\begin{figure}[h]
\centering
\includegraphics[width=0.6\linewidth]{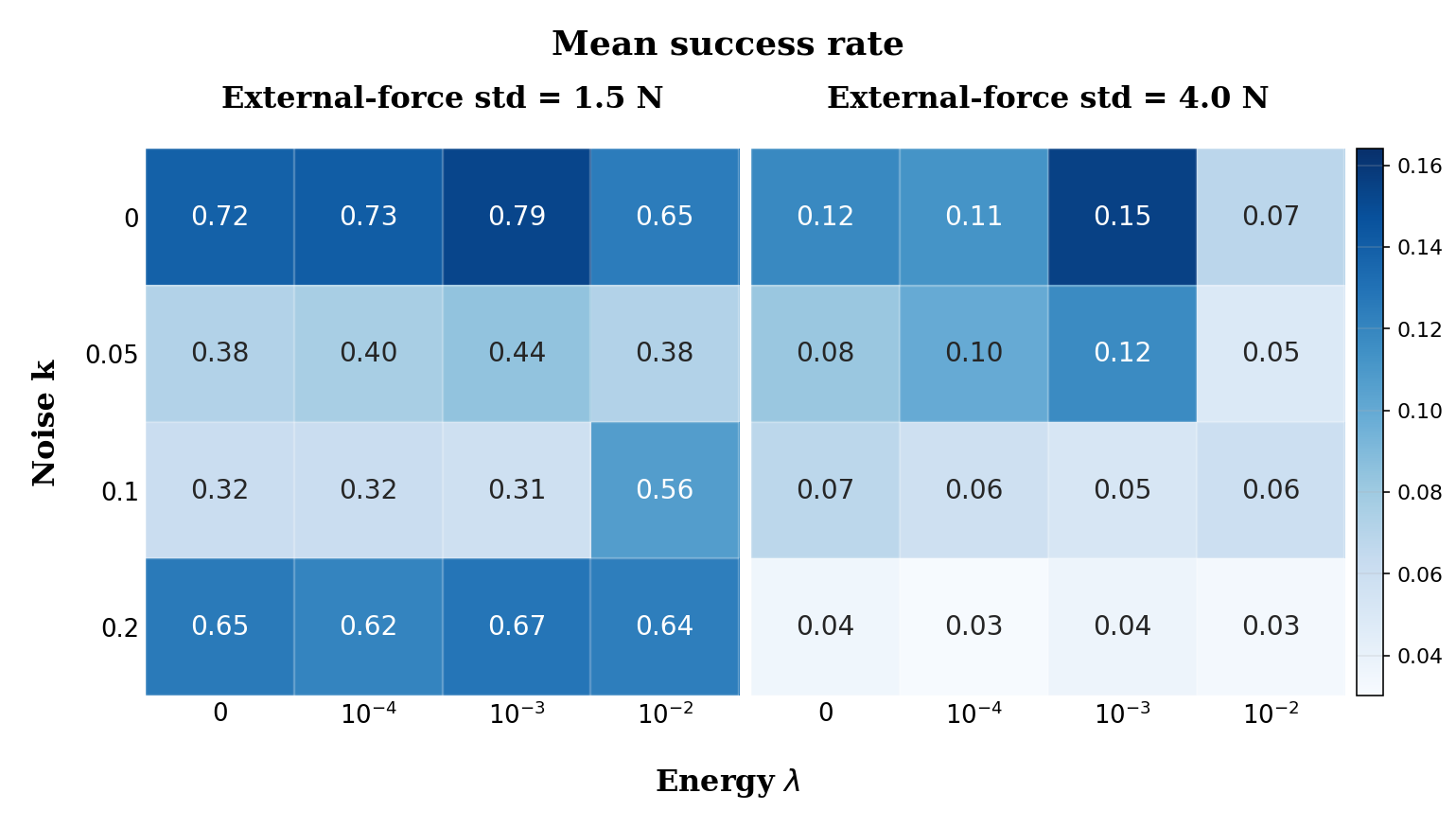}
\caption{Perturbation-strength diagnostic for the monolithic baseline. Heatmaps show mean success rate across motor-noise and energy-penalty conditions across ten random seeds when the Gaussian external-force standard deviation is increased to $1.5\,\mathrm{N}$ per axis (left) and $4.0\,\mathrm{N}$ per axis (right).}
\label{fig:app_high_noise_perturbation}
\end{figure}

Figure~\ref{fig:app_high_noise_perturbation} tests whether this rebound is robust to stronger holding disturbances by increasing the per-axis standard deviation of the Gaussian external force. Under a moderate increase to $1.5\,\mathrm{N}$, the high-noise rebound remains visible, consistent with the low-output regime remaining viable when the disturbance distribution is only moderately stronger than in training. Under a severe increase to $4.0\,\mathrm{N}$, the same high-noise region collapses toward floor performance. This argues against interpreting the rebound as genuinely improved active stabilisation.

\begin{figure}[h]
\centering
\includegraphics[width=0.6\linewidth]{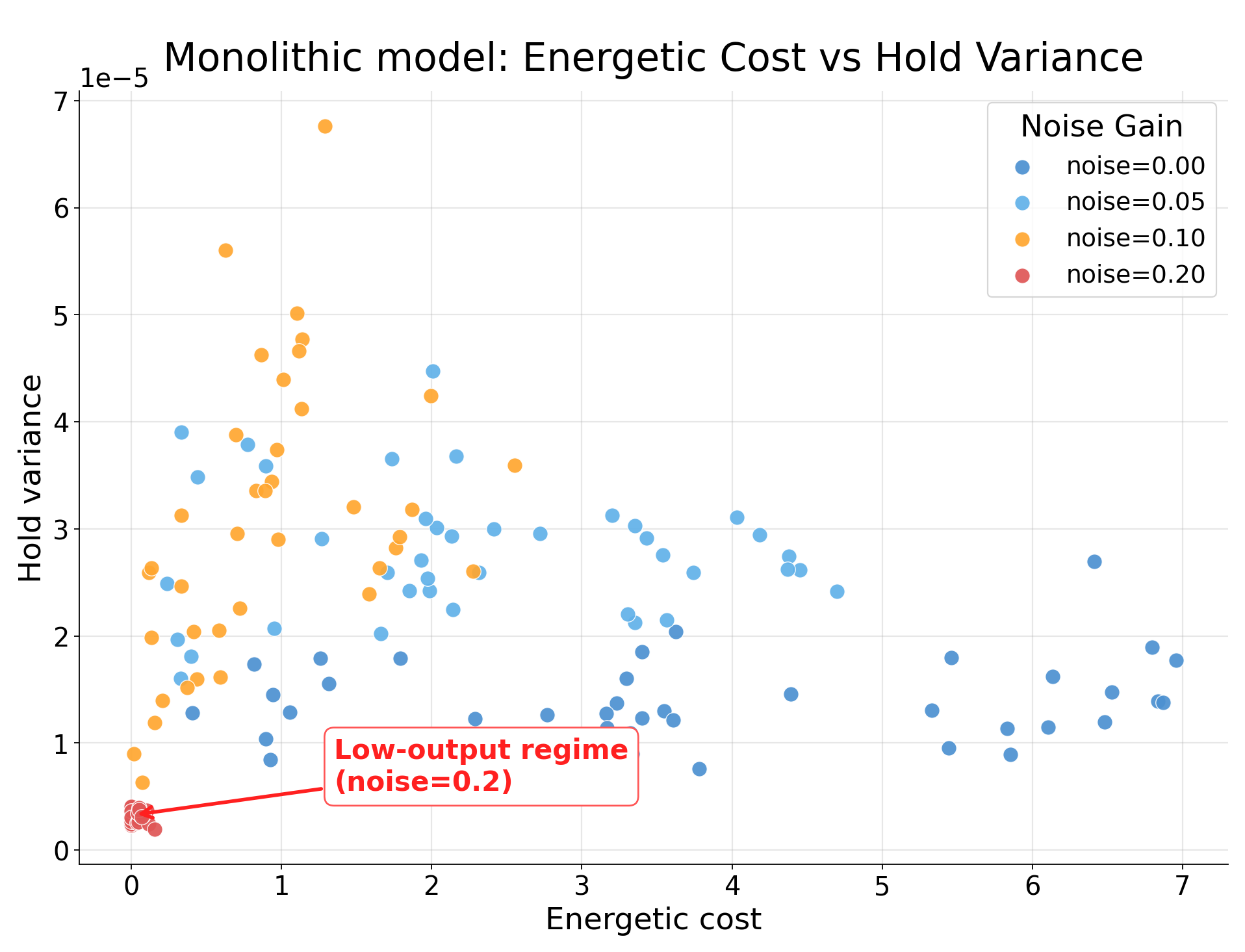}
\caption{Energy--variance diagnostic for the monolithic baseline across motor-noise conditions. Points show energetic cost and hold variance, grouped by motor-noise level $k$.}
\label{fig:app_high_noise_energy}
\end{figure}

Finally, the energy--variance diagnostic in Figure~\ref{fig:app_high_noise_energy} supports the low-output interpretation. At $k=0.2$, the points cluster near very low energetic cost and low hold variance, indicating that a trained model can satisfy the hold threshold with minimal commanded activation. The ambiguity is therefore behavioural: a low-output trained model can appear successful when the disturbance is weak enough, even though it does not provide robust active stabilisation under stronger disturbances.

Taken together, these diagnostics show why high-noise success rate should not be interpreted as evidence that trained models using the monolithic controller architecture learned better active muscle control. The rebound is hold-driven, depends on the original disturbance scale, and is better understood as a low-output holding regime exposed by signal-dependent noise. The stronger-disturbance check shows that this regime is fragile, supporting the main-text use of actuation-level metrics to interpret high-noise success.
\end{document}